\documentclass[journal]{IEEEtran}

\usepackage{graphicx}
\usepackage{algorithm}
\usepackage{algorithmic}
\usepackage{amsmath,amssymb,amsfonts,scalerel}
\usepackage{booktabs}

\newcommand{\bettercircled}[1]{\raisebox{0.15ex}{\textcircled{\raisebox{-0.15ex}{#1}}}}
\ifCLASSINFOpdf
\else
\fi
\begin{document}
\title{FedDAG: Federated Domain Adversarial Generation Towards Generalizable Medical Image Analysis}

\author{Haoxuan Che\textsuperscript{$*$}, \IEEEmembership{Student Member, IEEE}, Yifei Wu\textsuperscript{$*$}, Haibo Jin, Yong Xia\textsuperscript{$\dagger$}, \IEEEmembership{Senior Member, IEEE}, and Hao Chen\textsuperscript{$\dagger$}, \IEEEmembership{Senior Member, IEEE}
\thanks{This work was supported by the National Natural Science Foundation of China (No. 62202403), Hong Kong Innovation and Technology Fund (Project No. MHP/002/22) and Shenzhen Science and Technology Innovation Committee Fund (Project No. SGDX20210823103201011).}
\thanks{H. Che and H. Jin are with the Department of Computer Science and Engineering, at the Hong Kong University of Science and Technology University, Hong Kong SAR, China. (e-mail: hche@cse.ust.hk, hjinag@connect.ust.hk).}
\thanks{Y. Wu and Y. Xia are with the National Engineering Laboratory for Integrated Aero-Space-Ground-Ocean Big Data Application Technology, School of Computer Science and Engineering, Northwestern Polytechnical University, Xi’an 710072, China. (e-mail: yfwu@mail.nwpu.edu.cn; yxia@nwpu.edu.cn).}
\thanks{H. Chen is with the Department of Computer Science and Engineering, Department of Chemical and Biological Engineering and Division of Life Science, Hong Kong University of Science and Technology, Hong Kong, China (e-mail: jhc@cse.ust.hk).}
\thanks{\textsuperscript{$*$} indicates that they contributed equally to this work.}
\thanks{\textsuperscript{$\dagger$} indicates corresponding author.}
}

\maketitle

\begin{abstract}
Federated domain generalization aims to train a global model from multiple source domains and ensure its generalization ability to unseen target domains.
{Due to the target domain being with unknown domain shifts, attempting to approximate these gaps by source domains may be the key to improving model generalization capability.}
Existing works mainly focus on sharing and recombining local domain-specific attributes to increase data diversity and simulate potential domain shifts.
{However, these methods may be insufficient since only the local attribute recombination can be hard to touch the out-of-distribution of global data.}
In this paper, we propose a simple-yet-efficient framework named Federated Domain Adversarial Generation (FedDAG).
{It aims to simulate the domain shift and improve the model generalization by adversarially generating novel domains different from local and global source domains.}
Specifically, it generates novel-style images by maximizing the instance-level feature discrepancy between original and generated images and trains a generalizable task model by minimizing their feature discrepancy.
{Further, we observed that FedDAG could cause different performance improvements for local models.
It may be due to inherent data isolation and heterogeneity among clients, exacerbating the imbalance in their generalization contributions to the global model.}
{Ignoring this imbalance can lead the global model's generalization ability to be sub-optimal, further limiting the novel domain generation procedure. }
Thus, to mitigate this imbalance, FedDAG hierarchically aggregates local models at the within-client and across-client levels by using the sharpness concept to evaluate client model generalization contributions.
{Extensive experiments across four medical benchmarks demonstrate FedDAG's ability to enhance generalization in federated medical scenarios.}
\end{abstract}


\begin{figure}[!tbp]
    \centering
    \includegraphics[width=1\columnwidth]{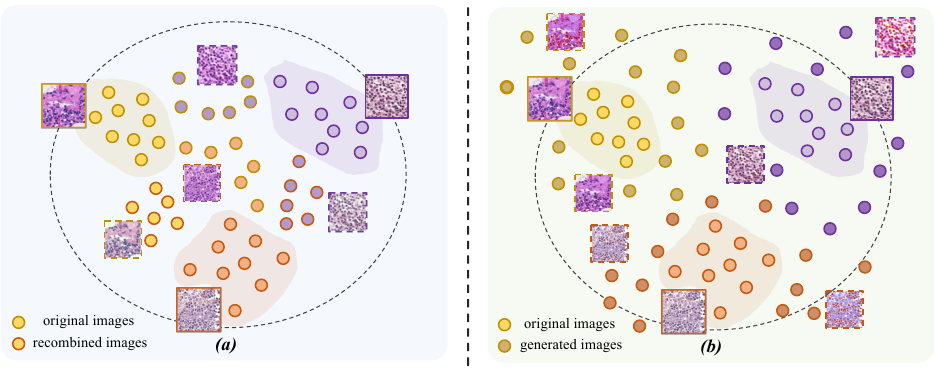}
    \caption{The potential solutions of approximating domain shifts in federated scenarios: (a) sharing and recombining local domain-specific attributes and (b) adversarial novel domain generation.}
    \vspace{-2mm}
    \label{fig1v}
\end{figure}

\IEEEpeerreviewmaketitle

\section{Introduction}
\IEEEPARstart{W}{ith} the continuous advancement of medical research and clinical practice, the medical field generates substantial data \cite{tang2024deep, ziqi2022using}.
However, these data are often scattered across different healthcare institutions. 
{Due to legal and regulatory restrictions, centralizing these data for training is probably infeasible \cite{rieke2020future, regulation2016regulation}.}
{Federated Learning (FL) \cite{mcmahan2017fedavg} enables multiple clients to collaborate on training models without data centralizing or sharing, ensuring that patient privacy is protected throughout the process.}
However, the existing methods of FL are sensitive to domain shifts, which occur when the target domain has a different data distribution than the source domains \cite{guo2023out}.
Neglecting the influence of domain shifts may limit the applications of FL models in real-world scenarios \cite{cao2023anomaly}.
{For instance, when a healthcare institution outside the federation aims to use the model trained by multiple other institutions inside the federation, the model often performs poorly due to the significant domain shifts \cite{liu2021feddg, dai2024deep, jin2023udaald}.}

Recently, Federated Domain Generalization (FedDG) has emerged as a new focus \cite{liu2021feddg}.
Unlike the setting of Domain Generalization (DG), FedDG aims to train a model from multiple source domains without data centralization and enable the model to generalize well on the unseen target domains \cite{liu2021feddg}. 
To achieve this generalization objective, various pioneered methods have been proposed \cite{liu2021feddg,chen2023ccst,zhou2023fedinb,park2024stablefdg,xu2023federated,nguyen2022fedsr,xu2022closing, zhang2023fedadg,zhang2023fedga,yuan2023csac,jiang2023iop, lai2023domain, le2024efficiently, guan2023rfdg, chen2023fraug,chen2024g2g,tuopu,DiPrompT}.
{For example, some existing methods primarily focus on model structure and aggregation strategies, aiming to effectively improve the generalization ability of the global model \cite{nguyen2022fedsr,xu2022closing, zhang2023fedadg,zhang2023fedga,yuan2023csac,jiang2023iop, lai2023domain, le2024efficiently, guan2023rfdg, chen2023fraug,chen2024g2g}. }
{Although these methods typically exhibit good generalization ability on known source domains, their performance on unseen target domains is often sub-optimal when significant domain shift occurs \cite{wang2023dafkd}.}
Other methods focus on increasing data diversity and simulating potential domain gaps by sharing and recombining local domain-specific attributes from different clients \cite{liu2021feddg,chen2023ccst,zhou2023fedinb,park2024stablefdg,xu2023federated}, such as the distribution of images and features.
{However, these domain attributes contain local private information.}
{Although it is challenging to reconstruct the training data from such attributes, their leakage remains prohibited due to regulatory constraints \cite{regulation2016regulation}.}
Moreover, recombining local domain attributes merely expands the local data yet still within the global distribution rather than touching the out-of-distribution of global data, as shown in Fig. \ref{fig1v} (a). This implies the difficulty for these methods to approximate the domain shifts, which limits their generalization capability on unseen target domains.

Novel domain generation is a promising solution to approximate the domain shifts and improve the model generalization ability to unseen target domains \cite{zhou2020ltnd,zhou2020deep,yang2021adversarial,wang2022learning, wgan, karras2019style, ho2020denoising, nichol2021improved, stabledif, zhang2024unpaired}.
{These methods can generate novel-style images with significant visual differences compared to source domains, through domain-level adversarial generation or the diffusion model with domain-related prompts.}
{The distribution of these generated images could differ from not only the local data distribution but also the global data distribution, as shown in Fig. \ref{fig1v} (b)}.
They expand the training data distribution, which can reflect domain shifts between target and source domains. 
{Therefore, effectively leveraging these generated images can improve model generalization capability.}
However, introducing novel domain generation into federated medical scenarios is non-trivial and faces challenges from two aspects.
{\textit{As for medical scenarios}, the gaps among the different clients are more ambiguous than natural scenarios \cite{che2023dgdr, che2022learning, zhou2022domain}.}
Thus, it may be insufficient to perform effective novel domain generation by the conventional learning paradigm, \textit{i.e.}, adversarial generating via domain labels or domain-related prompts \cite{zhou2020ltnd,zhou2020deep}.
{Moreover, constraining the semantics is essential, which are susceptible in the generated process.}
{Uncontrolled generation may lead to the loss of some key semantics, significantly affecting the generalization performance of the model \cite{che2023iqad, konz2024effect, qian2023drac, jin2024rethinking}.}
\textit{As for federated scenarios}, conducting novel domain generation on different clients can cause divergent performance improvements of client models, due to the data isolation and heterogeneity \cite{yuan2021we}.
{It can exacerbate the generalization contribution imbalance of local models to the global model, neglecting such imbalance may lead the model generalization to be sub-optimal \cite{zhang2023fedga}.}
{As can be imagined, the sub-optimal model can limit generation effectiveness since it plays an important role in the adversarial generation paradigm.}

{This paper proposes Federated Domain Adversarial Generation (FedDAG), a simple-yet-efficient framework to address the above challenges and improve the model generalization ability by adversarially generating data different from global data distribution.}
Specifically, it performs adversarial generation on each client locally, aiming to generate novel-style images with similar semantics but different visual styles.
{To encourage the generation of images with novel styles beyond the global rather than merely local distribution, FedDAG shares generators from different clients.}
{Besides, to tackle the challenges raised by medical scenarios, FedDAG generates images by performing instance-level adversarial learning, bypassing the challenges caused by ambiguous labels in the medical scenarios.}
FedDAG maintains the semantics consistency of generated images by dual-level semantic constraints in the task and image pixels.
{As for the challenge of federated scenarios, FedDAG leverages the concept of sharpness to evaluate client model generalization contribution and conduct model aggregation hierarchically to mitigate the differences in generalization contributions exacerbated by local generation.}
There, the within-client level aggregation focuses on balancing model contributions from historical communication rounds within a client, and the across-client level aggregation aims to mitigate the imbalance of model contributions to the global model among clients. 
{This aggregation mechanism improves the global model generalization and further promotes the adversarial novel domain generation procedure.}
We highlight our contributions as follows:
\begin{itemize}
    \item To the best of our knowledge, this is the first work to introduce novel domain generation into federated medical scenarios to improve generalization on diagnosis.
    \item We proposed FedDAG to overcome challenges raised by federated and medical scenarios to approximate domain shifts and leverage them effectively.
    \item {We conducted extensive experiments on four classic medical diagnosis benchmarks of there imaging modalities, proving the generalization capability of FedDAG. }
\end{itemize}

\begin{figure*}[!htb]
    \centering
    \includegraphics[width=0.99\textwidth]{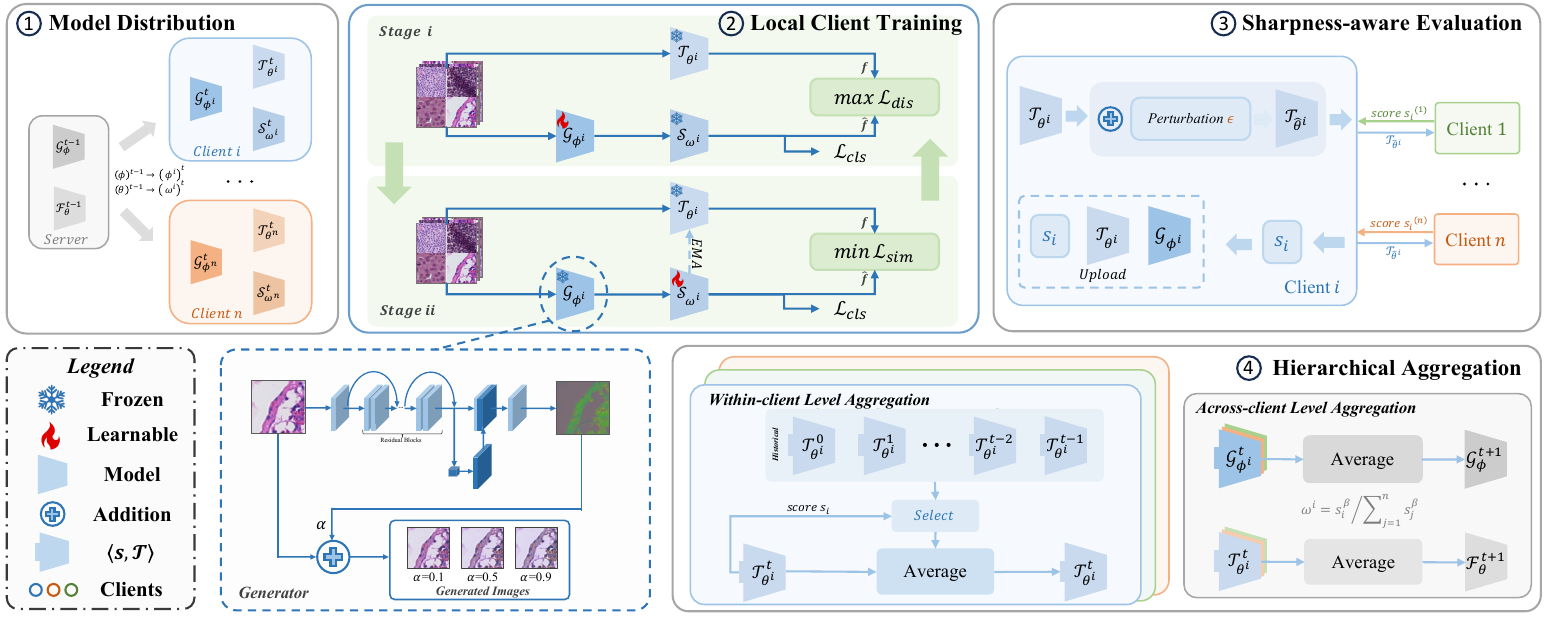}
    % \vspace{-2.5mm}
    \caption{An overview of FedDAG. It enhances the model's generalization by locally performing NDAG. 
    % Subsequently, sharpness-aware evaluation is employed to measure generalization contributions and further aggregate models.
    Subsequently, SHA is employed to mitigate the NDAG-exacerbated imbalance of generalization contributions and further promote the novel domain generation.
    }
    \label{Main_fig2}
    \vspace{-3mm}
\end{figure*}

\section{Related Work}

\subsection{Domain Generalization} 
Domain Generalization (DG) aims to train a model that can generalize under the domain shifts between the target and source domains by centralizing data from multiple data distributions \cite{zhou2022domain}.
{Some existing DG methods aim to learn domain-invariant representations by reducing the focus on style information of source data \cite{motiian2017unified, shao2019multi,dou2019domain}.}
For example, some methods utilize contrastive loss to mitigate the disparity between data from different domains  \cite{motiian2017unified, shao2019multi}, and some conduct meta-learning based on the potential domain shifts simulated by dividing the source data into disjoint meta-train and meta-test sets \cite{dou2019domain}.
{Moreover, some methods improve model generalization from the perspective of increasing data diversity \cite{zhou2020ltnd,zhou2020deep,yang2021adversarial,wang2022learning}.}
Representatively, DDAIG trains a generator to generate images that are recognizable by the task classifier but not by the domain classifier \cite{zhou2020deep}, while ATS maintains the difference between generated and source data based on a teacher-student paradigm, thereby increasing source data diversity \cite{yang2021adversarial}.
{From this perspective, generative adversarial networks (GANs) or diffusion methods can also be used to solve DG problems based on their effectiveness in creative generation \cite{wgan, karras2019style,ho2020denoising,nichol2021improved,stabledif, zhang2024unpaired}.}
{However, despite potentially improving generalization on the unseen target domain, these methods typically benefit from or necessitate data centralization during training, which is infeasible in FL due to privacy concerns.}
{Thus, although these DG methods can be performed without data centralization \cite{zhou2020deep, yang2021adversarial, wang2022learning, wgan, karras2019style, ho2020denoising, nichol2021improved, stabledif}, merely exploiting information from a single client leads the model to be sub-optimal \cite{xu2023federated}.}
{In contrast, this paper focuses on training a federated model with good generalization under data isolation, which is more challenging than centralized DG.}
Among these works, novel domain generation \cite{zhou2020ltnd,zhou2020deep,yang2021adversarial,wang2022learning} inspires FedDAG the most.
{However, FedDAG carefully considers the challenges posed by the federated medical scenarios and effectively addresses domain label ambiguity, semantic constraints, and imbalanced generalization contributions that previous methods cannot meet.}

% they are proposed for centralized natural scenarios \cite{zhou2020ltnd,zhou2020deep,yang2021adversarial,wang2022learning}. 

\subsection{Federated Domain Generalization} 
{FedDG is an emerging field that prioritizes generalization capability to the target domain while adhering to strict federated privacy constraints \cite{liu2021feddg}.}
{Different from only focusing on the inside-federation data distributions, FedDG also emphasizes improving generalization capability on unseen target domains \cite{huang2023FLG}.}
Recently, various pioneered methods have been proposed to focus on unseen domain shifts in FL scenarios \cite{liu2021feddg,chen2023ccst,zhou2023fedinb,park2024stablefdg,xu2023federated,nguyen2022fedsr,xu2022closing, zhang2023fedadg,zhang2023fedga,yuan2023csac,jiang2023iop,lai2023domain, le2024efficiently, guan2023rfdg, chen2023fraug, chen2024g2g,DiPrompT,tuopu}.
These methods can be broadly categorized into two categories. 
On the one hand, some existing methods focus on designing the model architecture and aggregation \cite{zhang2023fedadg,zhang2023fedga,nguyen2022fedsr,xu2022closing,yuan2023csac,jiang2023iop, lai2023domain, le2024efficiently, guan2023rfdg, chen2023fraug,chen2024g2g,tuopu}.
For example, FedGA focuses on designing the model aggregation strategy, which improves the global model generalization capability by dynamically calibrating the model weighting in the aggregation stage \cite{zhang2023fedga}.
% FedSR emphasizes the architecture design, which makes the model focus on learning domain-invariant representation information through setting a regularizer \cite{nguyen2022fedsr}.
{TFL utilizes a topological graph representing the relationships of different clients to infer client relationships, thereby updating the topological graph without leakaging privacy \cite{tuopu}.}
{However, due to domain gaps caused by federated data isolation and heterogeneity, local models of these methods often exhibit bias towards the local data distributions.
It may lead to the sub-optimal generalization capability of the aggregated model to unseen target domains \cite{dong2023no,miao2023fedseg}. }
{On the other hand, some methods improve model generalization capability by increasing data diversity and approximating unseen domain shifts \cite{liu2021feddg,xu2023federated,chen2023ccst,zhou2023fedinb, park2024stablefdg,DiPrompT}.}
% ,VHL,morafah2024stable,wu2024prompt,pmlr-v235-guo24c
Representatively, ELCFS \cite{liu2021feddg} and CCST \cite{chen2023ccst} share the local domain-specific attributes among different clients and recombine them to approximate the potential domain shifts.
{However, these types of data information may still be sensitive in FL scenarios, and sharing this information may violate federated privacy constraints \cite{regulation2016regulation}.}
{Additionally, these methods also cannot effectively approximate the unseen domain shifts since the sharing and recombination of local attributes do not obtain data outside the global distribution.}
{Recently, due to the success of generative methods, the GAN and diffusion-based methods have also been applied to federated scenarios due to their effective creative generation capability \cite{morafah2024stable, VHL, peng2019federated, pmlr-v235-guo24c,wu2024prompt,jin2023promptmrg}}
{For example, GenFedSD \cite{morafah2024stable} and VHL \cite{VHL} leverages Stable Diffusion \cite{stabledif} and StyleGAN \cite{karras2019style} in federated scenarios, respectively.}
{However, collecting large amounts of data for generative model training is difficult due to data scarcity and isolation in federal medical scenarios, thus limiting their effectiveness.}
{In contrast, FedDAG is designed for federated medical scenarios and approximates the unseen domain shifts by performing instance-level adversarial learning to generate perturbated images.
It bypasses the challenges caused by ambiguous labels in medical scenarios and demonstrates well-generalized performance even under a limited sample amount.}

\section{Methodology}

\subsection{Preliminaries}
% Preliminaries Setting

\subsubsection{Learning Objective}
We denote the set of source domains in this paper as $\mathcal{D}=\{D_1, D_2, ..., D_n\}$, with client $i$ holding a set of $N_i$ image and label pairs $D_i=\{(x_k^i,y_k^i)\}_{k=1}^{N^i}$, drawn from a data distribution $(\mathcal{X}^i, \mathcal{Y})$. 
This paper aims to leverage $n$ distributed client data to train a federated model $\mathcal{F}$ that can generalize well to the unseen target domain $\mathcal{U}$ by generating novel-style images and further leveraging these images.

\subsubsection{Overall Framework}
As shown in Fig. \ref{Main_fig2}, FedDAG follows a server-client architecture.
{It performs novel domain adversarial generation to generate images with similar semantics but different styles via maximizing instance-level feature discrepancy, and improves the task model generalization capability by leveraging these generated images via minimizing the discrepancy ($\bettercircled{2}$, Sec. \ref{NDG}).
Further, it leverages the concept of sharpness to evaluate the generalization contribution ($\bettercircled{3}$) and conduct model aggregation hierarchically ($\bettercircled{4}$), leading to obtaining a generalizable global model and further promoting the novel domain generation procedure (Sec. \ref{SHA}).}

\subsection{Novel Domain Adversarial Generation} \label{NDG}
{To tackle the challenge of applying novel domain generation in medical scenarios, we introduce two-stage novel domain adversarial generation (NDAG) in each local client training.}
{It first trains a generator to generate novel-style images by maximizing instance-level feature discrepancy from original and generated images while leveraging dual-level semantic constraints to ensure the semantics of generated images.}
{Then, it further leverages generated images to train a generalizable model by minimizing instance-level feature discrepancy from original and generated images.}
These two steps are executed in each mini-batch training, which can be viewed as a mutually adversarial process.
{FedDAG solves the challenges of domain label ambiguity and semantic sensitivity by repeating the above training process on the client side, effectively introducing novel domain generation into medical scenarios.}

\subsubsection{Instance-level Adversarial Generation Stage} \label{iag}
{NDAG utilizes instance-level adversarial learning to generate novel-style images and approximate domain shifts, thus bypassing the requirement of domain labels and avoiding the influence of their ambiguity.}
Specifically, each client has three local models: a local generator $\mathcal{G}_{\phi^i}$, a local teacher model $\mathcal{T}_{\theta^i}$, and a local student model $\mathcal{S}_{\omega^i}$. 
{During this stage, only $\mathcal{G}_{\phi^i}$ is allowed to update, while the $\mathcal{T}_{\theta^i}$ and $\mathcal{S}_{ \omega^i}$ are kept frozen.}
Given an image $x$ and its one-hot label $y$, NDAG trains $\mathcal{G}_{\phi^i}$ by exploiting the instance-level feature discrepancy between $x$ and its generated version $\hat{x}$. 
Once $\mathcal{G}_{\phi^i}$ generates $\hat{x}$, $\hat{x}$ is processed by $ \mathcal{S}_{\omega^i}$, \textit{i.e.}, $\hat{f} = \mathcal{S}_{\omega^i} (\hat{x})$. 
Simultaneously, the original image $x$ is input into $\mathcal{T}_{\theta^i}$, \textit{i.e.}, $f = \mathcal{T}_{\theta^i}(x)$. 
{As can be imagined that when there is no significant difference between $\mathcal{S}_{\omega^i}$ and $\mathcal{T}_{\theta^i}$, the feature discrepancy between $f$ and $\hat{f}$ highly correlated with the image visual difference between $x$ and generated $\hat{x}$. }
Thus, NDAG trains $\mathcal{G}_{\phi^i}$ to generate novel-style images, by maximizing the discrepancy loss $\mathcal{L}_{dis}$ between the normalized $\hat{f}$ and $f$, as
\begin{equation}
    \max_{\phi^i} \mathcal{L}_{dis}(f, \hat{f}) = \min_{} (
    \left\| \frac{f}{\|f\|_2} - \frac{\hat{f}}{\|\hat{f}\|_2} \right\|_2^2, m).
\label{max}
\end{equation}
Here, $ m $ is a hyperparameter to avoid the unlimited maximization of $\mathcal{L}_{dis}$, which is set to $0.1$ in our experiments.
{It can also be seen as a controlling factor for the difference between the generated images and the original images.}
Through the above process, the $ \mathcal{G}_{\phi^i} $ tackles the challenge of domain label ambiguity and generates images with novel styles by utilizing instance-level adversarial gradients.

\subsubsection{Dual-level Semantics Constraints} \label{dsc}
{Another essential design of NDAG is to constrain the semantics of susceptible medical images in the generation process, which requires preserving the original image information during the generation process.}
{Although the semantics changing of generated images may means achieve the significant domain shift approximation, this semantic changing may lead to difficulty in leveraging generated images.}
Therefore, NDAG performs a dual-level semantic constraint in the task and image pixels to avoid semantic changes in the generated process.
As for the image level, NDAG leverages a controllable image perturbation solution in the $\mathcal{G}_{\phi^i}$, represented as
\begin{equation}
    \hat{x} := x + \alpha \cdot \mathcal{G}_{\phi^i}(x),
\end{equation}
where $ \alpha $, lying within the range of $[0, 1]$, is designed to modulate the intensity of image perturbation. 
Compared to direct generation solutions \cite{yang2021adversarial,morafah2024stable, VHL, wu2024prompt}, this design enables the generation of novel-style images by adding perturbations and thus effectively preserving their original semantic content and reducing the demand for training data \cite{zhou2020deep}.
{As for the task level, NDAG train $\mathcal{G}_{\phi^i}$ to ensure that the generated image $ \hat{x} $ can be accurately classified by the $\mathcal{S}_{\omega^i}$, as}
\begin{equation}
    \min_{\phi^i} \mathcal{L}_{cls}(\hat{x}, y) = \text{CrossEntropy}(\mathcal{S}_{\omega^i}(\hat{x}), y).
\label{ce_g}
\end{equation}
{Through performing the dual-level semantic constraint, the semantics of the generated images can remain unchanged, which provides the foundation for learning the domain-invariant representation using these generated images.}

\subsubsection{Domain-invariant Representation Learning Stage} \label{drl}
After the generation, it is crucial to effectively leverage generated images.
{NDAG uses generated and original images to train $\mathcal{S}_{\omega^i} $ to resist the influence of perturbation, aiming to improve model generalization capability via learning domain-invariant features.}
Specifically, the $\mathcal{S}_{\omega^i}$ is activated to be trainable, while both the $ \mathcal{G}_{\phi^i} $ and $\mathcal{T}_{\theta^i}$ are kept frozen. 
{The idea is straightforward: when there is a significant difference between $\hat{x}$ and $x$ due to the presence of perturbations, if $\mathcal{S}_{\omega^i}$ can generate a feature representation similar to $\mathcal{T}_{\theta^i}$, it indicates that $\mathcal{S}_{\omega^i}$ possesses domain-invariant generalization capabilities. }
NDAG achieves this objective by minimizing the feature similarity loss $ \mathcal{L}_{sim}$ of normalized $\hat{f}$ and $f$, as
\begin{equation}
    \min_{\omega^i} \mathcal{L}_{sim}(f, \hat{f}) = \left\| \frac{f}{\|f\|_2} - \frac{\hat{f}}{\|\hat{f}\|_2} \right\|_2^2 .
\label{min}
\end{equation}
%- Such a minimization process guides $\mathcal{S}_{\omega^i}$ to resist image perturbations and thus enhances its capability to learn domain-invariant representation information.
Such a minimization guides $\mathcal{S}_{\omega^i}$ to resist image perturbations and thus enhances its capability to learn domain-invariant feature representations.
{Alongside, a task loss $\mathcal{L}_{cls}$, similar to Eq. \ref{ce_g}, is implemented to prevent $\mathcal{S}_{\omega^i}$ overfitting only to feature similarity while maintaining its Generalization performance.}
{Furthermore, to avoid the learning collapse arising from a significant knowledge gap between $\mathcal{T}_{\theta^i}$ and $\mathcal{S}_{\omega^i}$.}
{NDAG follows the teacher-student paradigm \cite{yang2021adversarial}, the Exponential Moving Average (EMA) \cite{tarvainen2017meantea} is employed to update $\mathcal{T}_{\theta^i}$ by progressive distillation of knowledge from $\mathcal{S}_{\omega^i}$.}

\begin{algorithm}
    \caption{Training of the $i$-th client in the $t$-th round} 
    \textbf{Input:} Local data $\mathcal{D}_{i}$,  student model $\mathcal{S}_{\omega^i}^{t}$, local generator $\mathcal{G}_{\phi_i}^{t}$ 
    \textbf{Output:} Local model $\mathcal{T}_{\theta^i}^{t}$ and local generator $\mathcal{G}_{\phi^i}^t$  
    
    \begin{algorithmic}[1]  
        \FOR {each mini-batch in $\mathcal{D}_{i}$}
            \STATE Generate novel style images (Sec. \ref{dsc})
            \STATE Optimize $\mathcal{G}_{\phi^i}^t$ via $\min_{\phi^i} (\mathcal{L}_{cls} -\mathcal{L}_{dis})$ (Sec. \ref{iag})
            \STATE Optimize $\mathcal{S}_{\omega^i}^t$ via $\min_{\omega^i} (\mathcal{L}_{cls}+\mathcal{L}_{sim})$ (Sec. \ref{drl})
            \STATE Update $\mathcal{T}_{\theta^i}^t$ with $\mathcal{S}_{\omega^i}^t$ via EMA
        \ENDFOR
        \STATE Upload $\mathcal{T}_{\theta^i}^t$ and $\mathcal{G}_{\phi^i}^t$ to the server
    \end{algorithmic}
    \label{Client}
\end{algorithm}

In summary, FedDAG employs NDAG on each client, enabling $\mathcal{G}_{\phi^i}$ to generate images with similar semantics but different styles.
{It further leverages these images to train the student model $\mathcal{S}_{\omega^i} $, aiming to learn domain invariant representation.}
{FedDAG deploy the teacher-student paradigm} 
{to balance the knowledge between $\mathcal{S}_ {\omega^i}$ and $\mathcal{T}_{\theta^i}$, and make the $\mathcal{T}_{\theta^i}$ as the final task model.}
The whole training process in local clients is described at Alg. \ref{Client}.

\subsection{Sharpness-aware Hierarchical Aggregation}\label{SHA}
{To mitigate the NDAG-exacerbated imbalance of generalization contributions and further promote the novel domain generation procedure, we propose Sharpness-aware Hierarchical Aggregation (SHA).}
{Through the evaluation of generalization contributions based on the concept of sharpness, each task model $\mathcal{T}_{{\theta}^i}$ receives a corresponding generalization score $s_i$ that reflects its generalization capability.}
Based on these scores, SHA performs hierarchical aggregation to mitigate the imbalance within and across clients, thereby improving the global model's generalization capability.
{It can be foreseen that due to the improved generalization ability of the global model, a better teacher model will be used to the next round of model distribution, which can further promote domain adversarial generation on the client side.}

\subsubsection{Sharpness-aware Model Evaluation} \label{SAE}
{SHA first evaluates the generalization contributions by the concept of sharpness \cite{foret2021sam, DFedSAM, FedLESAM, WIMA}.}
{Different from other evaluation methods, the sharpness-aware evaluation assesses the model's generalization ability toward unseen target domains by focusing on the flatness in loss landscapes.}
% {Following this concept, FedSAM \cite{Caldarola2022fedsam} obtains a model close to the flat minima on each client by densely model averaging, and believes that they can bring better generalization ability to the global model.}
{SHA quantifies the generalization ability of the model and performs densely weighted aggregation by simply adding small perturbations and observing the degree of change in generalization performance.}
Specifically, each client computes a model perturbation $\epsilon$ based on the dual norm of the last batch gradient $\nabla_{\omega^i} (\mathcal{L}_{cls} + \mathcal{L}_{sim})$, following \cite{foret2021sam}. 
The perturbed $\mathcal{T}_{\hat{\theta}^i}$ is obtained by adding $\epsilon$ to its parameters as $\hat{\theta}^i:= \theta^i + \rho \cdot \epsilon$, where $\rho$ within $[0, 1]$ controls the perturbation intensity.
{Further, the generalization scores $s_i$ of $\mathcal{T}_{\hat{\theta}^i}$ is evaluated by other clients against their local validation sets $\hat{D}_j$ as}
\begin{equation}
    s_i =  1 /  \big( \sum_{j=1}^n \mathbb{E}_{(x^\prime,y^\prime) \in \hat{D}_j} \big[ \mathcal{L}_{cls}(\mathcal{T}_{\hat{\theta}^i}(x^\prime),y^\prime) \big]  \big).
\end{equation}
{Through the sharpness-aware evaluation, each pair of $\mathcal{T}_{\theta^i}^t$ and $\mathcal{G}_{\phi^i}$ receives a generalization score $s_i$, which reflects the model generalization contribution and is used for hierarchical model aggregation.}
Then, this score with the corresponding $\mathcal{T}_{\theta^i}^t$ and $\mathcal{G}_{\phi^i}$ will be uploaded to the server.

\begin{algorithm}
    \caption{Evaluation and Aggregation for the $t$-th round}
    \textbf{Input:} Client number $n$, $\mathcal{T}_{\theta^i}^t$ and, $\mathcal{G}_{\phi^i}^t$ trained by each client \\
    \textbf{Output:} Global model $\mathcal{F}_{\theta}^{t+1}$, global generator $\mathcal{G}_{\phi}^{t+1}$
    
    \begin{algorithmic}[1]
        \FOR{each client $i$ \textbf{in parallel}}  
            
            \STATE Evaluate $\mathcal{T}_{\hat{\theta^i}}^t$ to obtain $s_i^t$ (Sec. \ref{SAE})
            \STATE Identifies $k$ models with scores above $s_i^t$ (Sec. \ref{WA})
            \STATE Average $\{\mathcal{T}_{\theta^i}^h, s_i^h\}_{h=1}^{k}$ to $\mathcal{T}_{\theta^i}^t$ and $s_i^t$ (Sec. \ref{WA})
        \ENDFOR
        \STATE Obtain $w^i$ via $\lbrace {s_i^t} \rbrace_{i=1}^N$ 
        \STATE Aggregate $\lbrace{\mathcal{T}_{\theta^i}^t}\rbrace_{i=1}^n$ via $w^i$ to $\mathcal{F}_{\theta}^{t+1}$ (Sec. \ref{AA})
        \STATE Aggregate ${\lbrace\mathcal{G}_{\omega^i}^t}\rbrace_{i=1}^n$ via $w^i$ to $\mathcal{G}_{\phi}^{t+1}$   (Sec. \ref{AA})
        % \STATE Distribute $\mathcal{F}_{\theta}^{t+1}$ and $\mathcal{G}_{\phi}^{t+1}$ to local clients
    \end{algorithmic}
    \label{Server}
\end{algorithm}

\subsubsection{Across-client Weighted Aggregation} \label{AA}
{As previously analyzed, NDAG may exacerbate the generalization contribution imbalance among clients.}
{Due to data isolation and heterogeneity, it can lead to different improvements in model generalization ability.}
Thus, SHA performs weighted aggregation among clients based on generalization scores to mitigate such imbalances and obtain generalizable global models.
Specifically, the server aggregates $\mathcal{T}_{\theta^i}^t$ and $\mathcal{G}_{\phi^i}^t$ based on the generalization score $s_i$ from different clients by using an adaptive soft-balancing weight \cite{che2023dgdr} as
\begin{equation}
    w^i = s_i^\beta / (\sum^n_{j=1} s_j^\beta), \quad \theta := \sum_{i=1}^n w^i \cdot \theta^i, \quad \phi := \sum_{i=1}^n w^i \cdot \phi^i,
\end{equation}
{where $\beta$ is related to the calculation of model aggregation weights $w^i$, which represents the aggregation weight of models from client $i$.}
By weighing the model aggregation, SHA mitigates the imbalances of generalization contribution among clients.
{In summary, the aggregation of $ \mathcal{G}_{\phi^i}^t $ is aimed to encourages the generation of novel-style images whose styles differ from the global distribution.}

\subsubsection{Within-client Densely Aggregation} \label{WA}
{As can be imagined, the generator continually generates different style images across local training rounds, and thus, corresponding task models in historical rounds may also contain diverse generalization knowledge.}
Therefore, merely aggregating the latest round models from each client may be insufficient.
To effectively leverage the diverse generalization knowledge in historical rounds, SHA balances generalization contributions from different communication rounds for each client before aggregating across clients.
{Therefore, inspired by seeking flat minima via model averaging \cite{cha2021swad, WIMA}, SHA aggregate models from historical training rounds within the same client to obtain the generalizable model.}
{Specifically, SHA retains the latest $k$ local models with scores exceeding $s_i^t$, integrates them with $\mathcal{T}_{\theta^i}^t$ via model averaging, and recalculates the average score.}
SHA mitigates the imbalance from different communication rounds for each client by averaging aggregation local model $\mathcal{T}_{\theta^i}^t$, thereby fully utilizing the generation effects of the NDAG.

In conclusion, SHA mitigates the NDAG-exacerbated imbalance of generalization contribution by evaluating the generalization contributions based on the concept of sharpness and aggregating models within and across clients hierarchically.
{By utilizing SHA, the model with more generalizable than simple average can be obtained, which further promotes the adversarial generation process of NDAG in the next round.}
The aggregation process is shown as Alg. \ref{Server}.

\subsection{Federated Domain Adversarial Generation}
FedDAG can be divided into four stages: model initialization, model distribution, local training and uploading (Alg. \ref{Client}), model evaluation and aggregation  (Alg. \ref{Server}).
\textit{Model Initialization:} As the same as FedAvg, FedDAG initializes global models on the server before the first model distribution.
\textit{Model Distribution}: FedDAG distributes $\mathcal{G}_{\phi}$ and $\mathcal{F}_{\theta}$ to local clients as $\mathcal{G}_{\phi^i}$ and $ \mathcal{S}_{\omega^i}$.
\textit{Local Training and Uploading}: FedDAG performs NDAG in each client to train $\mathcal{T}_{\theta^i}$ and $\mathcal{G}_{\phi^i}$ locally, and after the training, the $\mathcal{T}_{\theta^i}$ and $\mathcal{G}_{\phi^i}$ will be uploaded to the server.
{\textit{Model Evaluation and Aggregation}: FedDAG evaluates the generalization capability of $\mathcal{T}_{\theta^i}$ and $\mathcal{G}_{\phi^i}$ based on the concept of flat minima, and then it hierarchically aggregates these models to obtain $\mathcal{F}_{\theta}$ and $\mathcal{G}_{\phi}$ based on this evaluation results.}
Then, the aggregated models are prepared for the next round of model distribution.
As can be seen, different from standard distribution and uploading, FedDAG introduces a chain: it distributes the $\mathcal{F}_{\theta}$ to the $\mathcal{S}_{\omega^i}$, and further the client transfers the knowledge of $\mathcal{S}_{\omega^i}$ to $\mathcal{T}_{{\theta}^i}$ during training and finally uploads $\mathcal{T}_{{\theta}^i}$ to obtain $\mathcal{F}_{\theta}$.
{Such a design aims to maintain the proper knowledge difference between the $\mathcal{T}_{{\theta}^i}$ and $ \mathcal{S}_{\omega^i}$, which can promote the generation of novel-style images.}

\begin{table*}[htb]
    \caption{Comparison results on the WILDS-Camelyon17. $\dagger$: w/ data info. sharing. $\ddagger$: w/ cross-client communication.}
    % \vspace{-2mm}
    \renewcommand\arraystretch{1.25}
    \centering
    \setlength{\tabcolsep}{2pt}
    \resizebox{1.0\textwidth}{!}{
        \scalebox{0.8}{
            \begin{tabular}{c|ccc|ccc|ccc|ccc|ccc|ccc} 
            \toprule
            % \hline
            % {}&\multicolumn{18}{c}{WILDS-Camelyon17}\\
            \hline
            Unseen Site &\multicolumn{3}{c}{Hospital 1}&\multicolumn{3}{c}{Hospital 2}&\multicolumn{3}{c}{Hospital 3}&\multicolumn{3}{c}{Hospital 4}&\multicolumn{3}{c}{Hospital 5}&\multicolumn{3}{c}{AVG} \\
            \hline      
            Metrics  & AUC & F1 & ACC & AUC & F1 & ACC & AUC & F1 & ACC & AUC & F1 & ACC & AUC & F1 & ACC & AUC & F1 & ACC \\
            \hline
            FedAvg \cite{mcmahan2017fedavg} & $94.2_{\pm{0.2}}$ & $89.4_{\pm{0.4}}$ & $91.4_{\pm{0.4}}$ & $90.7_{\pm{0.0}}$ & $78.8_{\pm{0.8}}$ & $80.3_{\pm{0.9}}$ & $89.0_{\pm{0.2}}$ & $71.2_{\pm{0.4}}$ & $71.4_{\pm{0.4}}$ & $92.3_{\pm{0.3}}$ & $80.4_{\pm{0.8}}$ & $81.8_{\pm{0.8}}$ & $87.9_{\pm{0.0}}$ & $77.9_{\pm{0.1}}$ & $79.6_{\pm{0.2}}$ & $90.8_{\pm{0.1}}$ & $79.5_{\pm{0.5}}$ & $80.9_{\pm{0.5}}$  \\
            \hline
            FedBN \cite{li2021fedbn} & $97.2_{\pm{0.2}}$ & $89.3_{\pm{1.1}}$ & $89.2_{\pm{1.2}}$ & $95.1_{\pm{0.5}}$ & $78.3_{\pm{2.7}}$ & $77.7_{\pm{3.3}}$ & $92.7_{\pm{0.4}}$ & $82.3_{\pm{2.1}}$ & $81.8_{\pm{2.2}}$ & $95.5_{\pm{0.6}}$ & $82.9_{\pm{1.3}}$ & $82.4_{\pm{1.4}}$ & $88.7_{\pm{0.7}}$ & $75.6_{\pm{3.1}}$ & $74.7_{\pm{3.6}}$ & $93.8_{\pm{0.5}}$ & $81.7_{\pm{2.1}}$ & $81.2_{\pm{2.3}}$  \\
            FedProx \cite{li2020federated} & $97.0_{\pm{0.2}}$ & $87.3_{\pm{2.5}}$ & $88.4_{\pm{2.5}}$ & $93.8_{\pm{0.5}}$ & $78.6_{\pm{2.1}}$ & $77.7_{\pm{2.3}}$ & $92.6_{\pm{1.6}}$ & $80.8_{\pm{3.0}}$ & $80.3_{\pm{3.2}}$ & $95.4_{\pm{0.6}}$ & $82.7_{\pm{1.7}}$ & $82.2_{\pm{1.8}}$ & $88.5_{\pm{1.5}}$ & $77.5_{\pm{2.3}}$ & $78.8_{\pm{1.9}}$ & $93.5_{\pm{0.9}}$ & $81.4_{\pm{2.3}}$ & $82.3_{\pm{2.3}}$ \\
            FCCL \cite{FCCL_CVPR22} & $93.3_{\pm{2.7}}$ & $82.7_{\pm{1.6}}$ & $81.5_{\pm{3.9}}$ & $94.7_{\pm{2.6}}$ & $76.8_{\pm{0.2}}$ & $80.6_{\pm{0.5}}$ & $90.7_{\pm{2.1}}$ & $87.4_{\pm{0.3}}$ & $87.0_{\pm{0.4}}$ & $91.6_{\pm{2.5}}$ & $75.8_{\pm{2.8}}$ & $79.2_{\pm{5.1}}$ & $88.4_{\pm{4.9}}$ & $74.7_{\pm{6.1}}$ & $75.8_{\pm{4.1}}$ & $91.7_{\pm{3.0}}$ & $79.5_{\pm{2.2}}$ & $80.8_{\pm{2.8}}$ \\
            Per-FAvg \cite{fallah2020personalized} & $95.8_{\pm{0.6}}$ & $81.9_{\pm{1.5}}$ & $84.6_{\pm{2.2}}$ & $93.7_{\pm{0.4}}$ & $77.8_{\pm{2.1}}$ & $82.1_{\pm{0.5}}$ & $90.8_{\pm{1.2}}$ & $81.9_{\pm{1.6}}$ & $83.3_{\pm{2.1}}$ & $91.7_{\pm{2.1}}$ & $78.2_{\pm{2.7}}$ & $78.6_{\pm{3.3}}$ & $86.1_{\pm{3.0}}$ & $73.6_{\pm{2.6}}$ & $74.6_{\pm{1.7}}$ & $91.6_{\pm{1.5}}$ & $78.7_{\pm{2.1}}$ & $80.6_{\pm{1.7}}$ \\
            \hline
            CCST$\dagger$ \cite{chen2023ccst} & $\underline{97.3}_{\pm{0.3}}$ & $90.4_{\pm{0.5}}$ & $90.4_{\pm{0.5}}$ & $\underline{95.6}_{\pm{0.2}}$ & ${85.6}_{\pm{2.2}}$ & ${85.8}_{\pm{1.9}}$ & ${93.5}_{\pm{0.4}}$ & ${86.5}_{\pm{2.5}}$ & ${86.2}_{\pm{2.3}}$ & $94.6_{\pm{0.1}}$ & ${83.8}_{\pm{2.5}}$ & ${82.5}_{\pm{2.3}}$ & $91.5_{\pm{0.8}}$ & $79.3_{\pm{2.1}}$ & $78.7_{\pm{2.2}}$ & $94.5_{\pm{0.4}}$ & $\underline{85.1}_{\pm{1.9}}$ & $\underline{84.7}_{\pm{1.8}}$ \\
            ELCFS$\dagger$ \cite{liu2021feddg} & ${96.3}_{\pm{0.8}}$ & $\textbf{93.4}_{\pm{0.9}}$ & $\textbf{93.4}_{\pm{1.0}}$ & $94.5_{\pm{0.7}}$ & $\underline{86.8}_{\pm{0.9}}$ & $\underline{86.6}_{\pm{1.0}}$ & $93.0_{\pm{1.5}}$ & $76.2_{\pm{2.1}}$ & $75.1_{\pm{2.4}}$ & ${95.1}_{\pm{0.8}}$ & $82.1_{\pm{4.1}}$ & $81.5_{\pm{4.6}}$ & $\underline{95.3}_{\pm{0.9}}$ & $76.8_{\pm{1.4}}$ & $75.7_{\pm{1.6}}$ & $\underline{94.8}_{\pm{0.9}}$ & $83.1_{\pm{1.9}}$ & $82.5_{\pm{2.1}}$  \\
            FedGA$\ddagger$ \cite{zhang2023fedga} & $97.6_{\pm{0.1}}$ & $\underline{92.8}_{\pm{2.5}}$ & $\underline{92.9}_{\pm{3.0}}$ & $94.6_{\pm{0.1}}$ & $79.4_{\pm{2.6}}$ & $80.6_{\pm{3.0}}$ & $92.8_{\pm{0.1}}$ & $78.5_{\pm{3.3}}$ & $77.6_{\pm{3.8}}$ & $94.4_{\pm{0.5}}$ & $78.1_{\pm{3.2}}$ & $76.8_{\pm{3.8}}$ & $90.2_{\pm{1.7}}$ & $79.5_{\pm{2.4}}$ & $79.1_{\pm{2.7}}$ & $93.9_{\pm{0.9}}$ & $81.7_{\pm{2.8}}$ & $81.4_{\pm{3.3}}$ \\
            CSAC$\ddagger$ \cite{yuan2023csac} & $92.8_{\pm{0.1}}$ & $85.6_{\pm{0.7}}$ & $85.5_{\pm{0.7}}$ & $89.7_{\pm{0.6}}$ & $76.2_{\pm{1.6}}$ & $75.3_{\pm{1.9}}$ & $87.2_{\pm{2.1}}$ & $77.8_{\pm{2.1}}$ & $77.5_{\pm{2.3}}$ & $\underline{95.7}_{\pm{0.3}}$ &  $\textbf{86.7}_{\pm{0.3}}$ & $\textbf{85.6}_{\pm{0.3}}$ & $91.9_{\pm{2.3}}$ & $78.6_{\pm{3.4}}$ & $77.9_{\pm{4.1}}$ & $91.5_{\pm{1.1}}$ & $80.9_{\pm{1.6}}$ & $80.4_{\pm{1.8}}$ \\
            FedSR \cite{nguyen2022fedsr} & $96.2_{\pm{0.3}}$ & $73.4_{\pm{1.2}}$ & $72.7_{\pm{1.1}}$ & $94.1_{\pm{2.6}}$ & $82.3_{\pm{4.9}}$ & $82.8_{\pm{4.1}}$ & $91.3_{\pm{2.1}}$ & $81.1_{\pm{5.2}}$ & $79.4_{\pm{4.7}}$ &  $94.8_{\pm{0.4}}$ & $78.8_{\pm{1.7}}$ & $80.6_{\pm{2.0}}$ & ${93.3}_{\pm{2.2}}$ & ${74.7}_{\pm{5.1}}$ & ${74.4}_{\pm{5.3}}$  & $93.9_{\pm{1.5}}$ & $78.1_{\pm{3.6}}$ & $77.9_{\pm{3.4}}$ \\
            {G2G \cite{chen2024g2g}} & {$96.8_{\pm{0.1}}$} & {$89.1_{\pm{0.4}}$} & $89.2_{\pm{0.4}}$ & $94.0_{\pm{0.4}}$ & $69.4_{\pm{5.4}}$ & $71.7_{\pm{4.2}}$ & $93.6_{\pm{1.7}}$ & $76.1_{\pm{6.6}}$ & $77.6_{\pm{5.4}}$ & $92.8_{\pm{1.9}}$ & $73.8_{\pm{1.6}}$ & $74.9_{\pm{1.9}}$ & ${85.5}_{\pm{3.4}}$ & ${74.1}_{\pm{3.9}}$ & ${74.9}_{\pm{3.5}}$ & $92.5_{\pm{1.5}}$ & $76.5_{\pm{3.6}}$ & $77.7_{\pm{3.1}}$ \\

            \hline
            RSC \cite{huang2020self} & $93.2_{\pm{1.4}}$ & $80.1_{\pm{4.0}}$ & $80.4_{\pm{4.1}}$ & $93.2_{\pm{1.4}}$ & $81.3_{\pm{3.9}}$ & $80.5_{\pm{4.5}}$ & $\underline{93.7}_{\pm{0.1}}$ & $\textbf{89.1}_{\pm{0.5}}$ & $\textbf{89.1}_{\pm{0.5}}$ & $93.2_{\pm{0.5}}$ & $79.7_{\pm{1.5}}$ & $78.7_{\pm{1.7}}$ & ${93.4}_{\pm{1.5}}$ & $\textbf{87.4}_{\pm{3.4}}$ & $\textbf{87.3}_{\pm{3.5}}$ & ${93.3}_{\pm{1.0}}$ & $83.5_{\pm{2.7}}$ & $83.2_{\pm{2.7}}$ \\
            Mixstyle \cite{zhou2021domain} & $93.3_{\pm{0.8}}$ & $75.2_{\pm{2.7}}$ & $73.7_{\pm{3.3}}$ & $93.8_{\pm{0.4}}$ & $74.7_{\pm{2.2}}$ & $73.0_{\pm{2.7}}$ & $88.9_{\pm{3.1}}$ & $80.5_{\pm{3.8}}$ & $80.4_{\pm{3.9}}$ & $85.9_{\pm{3.8}}$&$ 69.9_{\pm{0.7}}$ & $67.6_{\pm{0.8}}$ & $\textbf{96.2}_{\pm{0.5}}$ & ${83.7}_{\pm{1.2}}$ & ${82.7}_{\pm{1.2}}$ & $91.6_{\pm{1.7}}$ & $76.8_{\pm{2.1}}$ & $75.5_{\pm{2.4}}$ \\
            ATS \cite{yang2021adversarial} & $94.1_{\pm{1.2}}$ & $89.5_{\pm{3.1}}$ & $89.5_{\pm{3.2}}$ & ${93.1}_{\pm{2.2}}$ & $75.8_{\pm{5.9}}$ & $76.7_{\pm{4.5}}$ & ${93.3}_{\pm{0.1}}$ & ${88.4}_{\pm{0.2}}$ & $\underline{88.4}_{\pm{0.2}}$ & ${94.7}_{\pm{0.4}}$ & $73.4_{\pm{2.9}}$ & $71.7_{\pm{4.2}}$ & $92.7_{\pm{1.7}}$ & $\underline{86.7}_{\pm{3.6}}$ & $\underline{86.6}_{\pm{3.7}}$ & $93.6_{\pm{1.1}}$ & $82.8_{\pm{3.1}}$ & $82.6_{\pm{3.2}}$  \\
            DDAIG \cite{zhou2020deep} & $96.8_{\pm{0.4}}$ & $87.3_{\pm{0.9}}$ & $87.5_{\pm{1.1}}$ & $94.1_{\pm{0.2}}$ & $80.4_{\pm{3.1}}$ & $79.8_{\pm{3.5}}$ & $93.3_{\pm{0.6}}$ & $82.8_{\pm{3.8}}$ & $82.4_{\pm{4.3}}$ & $94.7_{\pm{0.3}}$ & $81.6_{\pm{2.8}}$ & $80.9_{\pm{3.1}}$ & $89.6_{\pm{1.3}}$ & $78.3_{\pm{1.8}}$ & $77.9_{\pm{1.9}}$ & $93.7_{\pm{0.6}}$ & $82.1_{\pm{2.5}}$ & $81.7_{\pm{2.8}}$  \\
            \hline
            \textbf{FedDAG} & $\textbf{97.5}_{\pm{0.5}}$ & ${92.4}_{\pm{0.4}}$ & ${92.4}_{\pm{0.4}}$ & $\textbf{97.1}_{\pm{0.2}}$ & $\textbf{90.8}_{\pm{0.7}}$ & $\textbf{90.1}_{\pm{0.7}}$ & $\textbf{94.8}_{\pm{0.7}}$ & $\underline{88.6}_{\pm{3.1}}$ & ${87.6}_{\pm{3.3}}$ & $\textbf{96.5}_{\pm{0.4}}$ & $\underline{85.7}_{\pm{0.6}}$ & $\underline{85.4}_{\pm{0.7}}$ & $94.8_{\pm{0.5}}$ & $85.0_{\pm{2.1}}$ & $84.7_{\pm{2.3}}$ & $\textbf{96.1}_{\pm{0.5}}$ & $\textbf{88.5}_{\pm{1.4}}$ & $\textbf{88.1}_{\pm{1.5}}$  \\
            \hline
            \bottomrule
            \end{tabular}
        }
    }
    \label{Camelyon_exp}
    \vspace{-3mm}
\end{table*}

\begin{table*}[htb]
    \renewcommand\arraystretch{1.25}
    \centering
    \setlength{\tabcolsep}{2pt}
    \caption{Comparison results on the MIDOG2022. $\dagger$: w/ data info. sharing. $\ddagger$: w/ cross-client communication.}
    % \vspace{-2mm}
    \resizebox{1.0\textwidth}{!}{
        \scalebox{0.8}{
            \begin{tabular}{c|ccc|ccc|ccc|ccc|ccc|ccc} 
            \toprule
            % \hline
            % {}&\multicolumn{18}{c}{MIDOG2022}\\
            \hline
            Unseen Site &\multicolumn{3}{c}{Hospital 1}&\multicolumn{3}{c}{Hospital 2}&\multicolumn{3}{c}{Hospital 3}&\multicolumn{3}{c}{Hospital 4}&\multicolumn{3}{c}{Hospital 5}&\multicolumn{3}{c}{AVG} \\
            \hline      
            Metrics  & AUC & F1 & ACC & AUC & F1 & ACC & AUC & F1 & ACC & AUC & F1 & ACC & AUC & F1 & ACC & AUC & F1 & ACC \\
            \hline
            FedAvg \cite{mcmahan2017fedavg} & $81.3_{\pm{1.6}}$ & $74.0_{\pm{2.1}}$ & $73.9_{\pm{1.9}}$ & $80.4_{\pm{2.1}}$ & $70.8_{\pm{1.1}}$ & $70.8_{\pm{1.1}}$ & $62.1_{\pm{6.1}}$ & $57.9_{\pm{5.5}}$ & $49.1_{\pm{10.3}}$ & $81.8_{\pm{1.5}}$ & $72.4_{\pm{3.0}}$ & $72.8_{\pm{2.9}}$ & $76.9_{\pm{1.8}}$ & $76.1_{\pm{1.2}}$ & $75.1_{\pm{0.7}}$ & $76.5_{\pm{2.6}}$ & $70.2_{\pm{2.6}}$ & $68.3_{\pm{3.4}}$  \\
            \hline
            FedBN \cite{li2021fedbn} & $81.1_{\pm{0.5}}$ & $75.5_{\pm{0.7}}$ & $75.2_{\pm{0.4}}$ & $78.9_{\pm{0.4}}$ & $70.4_{\pm{0.7}}$ & $70.3_{\pm{0.8}}$ & $58.2_{\pm{1.0}}$ & $53.9_{\pm{0.3}}$ & $41.3_{\pm{1.5}}$ & $75.9_{\pm{5.6}}$ & $68.1_{\pm{6.6}}$ & $68.3_{\pm{6.6}}$ & ${78.3}_{\pm{1.7}}$ & $\textbf{77.9}_{\pm{1.2}}$ & $\underline{76.3}_{\pm{2.9}}$ & $74.5_{\pm{1.8}}$ & $69.2_{\pm{1.9}}$ & $66.3_{\pm{2.4}}$ \\
            FedProx \cite{li2020federated} & $79.8_{\pm{0.5}}$ & $72.6_{\pm{0.7}}$ & $70.7_{\pm{0.4}}$ & $78.6_{\pm{0.4}}$ & $70.9_{\pm{0.7}}$ & $70.5_{\pm{0.8}}$ & $70.4_{\pm{1.0}}$ & $65.1_{\pm{0.3}}$ & $\textbf{64.9}_{\pm{1.5}}$ & $\underline{82.9}_{\pm{5.6}}$ & $73.6_{\pm{6.6}}$ & $74.0_{\pm{6.6}}$ & $74.4_{\pm{1.7}}$ & $75.6_{\pm{1.2}}$ & $74.5_{\pm{2.9}}$ & $77.2_{\pm{1.8}}$ & $71.6_{\pm{1.9}}$ & $70.9_{\pm{2.4}}$ \\
            FCCL \cite{FCCL_CVPR22} & $76.3_{\pm{1.1}}$ & $71.8_{\pm{0.8}}$ & $70.5_{\pm{0.7}}$ & $69.4_{\pm{2.7}}$ & $63.8_{\pm{2.2}}$ & $63.5_{\pm{2.0}}$ & $65.5_{\pm{1.4}}$ & $60.5_{\pm{0.7}}$ & $58.7_{\pm{1.3}}$ & $68.1_{\pm{7.2}}$ & $60.4_{\pm{6.8}}$ & $59.6_{\pm{5.9}}$ & $74.0_{\pm{0.7}}$ & $70.5_{\pm{1.2}}$ & $71.4_{\pm{1.0}}$ & $70.7_{\pm{2.6}}$ & $65.4_{\pm{2.3}}$ & $64.8_{\pm{2.2}}$ \\
            Per-FAvg \cite{fallah2020personalized} & $77.5_{\pm{4.8}}$ & $75.2_{\pm{3.5}}$ & $74.3_{\pm{4.1}}$ & $67.4_{\pm{4.6}}$ & $66.2_{\pm{1.5}}$ & $59.5_{\pm{5.1}}$ & $66.4_{\pm{5.0}}$ & $62.8_{\pm{2.3}}$ & $60.7_{\pm{1.1}}$ & $67.7_{\pm{1.2}}$ & $65.6_{\pm{1.4}}$ & $62.5_{\pm{0.2}}$ & $70.0_{\pm{1.6}}$ & $60.3_{\pm{8.2}}$ & $60.0_{\pm{11.8}}$ & $69.8_{\pm{3.4}}$ & $66.0_{\pm{3.4}}$ & $63.4_{\pm{4.5}}$ \\
            \hline
            CCST$\dagger$ \cite{chen2023ccst} & $\underline{84.2}_{\pm{0.1}}$ & $\textbf{78.5}_{\pm{0.4}}$ & $\textbf{78.2}_{\pm{0.4}}$ & ${78.0}_{\pm{1.7}}$ & ${70.2}_{\pm{1.8}}$ & ${69.7}_{\pm{1.7}}$ & ${70.0}_{\pm{2.3}}$ & ${65.0}_{\pm{1.3}}$ & ${64.2}_{\pm{1.7}}$ & $81.9_{\pm{0.9}}$ & ${72.0}_{\pm{3.0}}$ & ${72.3}_{\pm{3.0}}$ & $76.5_{\pm{3.9}}$ & $75.8_{\pm{3.2}}$ & $74.4_{\pm{3.4}}$ & $\underline{78.1}_{\pm{1.8}}$ & ${72.3}_{\pm{1.9}}$ & ${71.8}_{\pm{2.0}}$ \\
            ELCFS$\dagger$ \cite{liu2021feddg} & ${83.7}_{\pm{1.3}}$ & ${76.7}_{\pm{0.6}}$ & ${76.0}_{\pm{1.0}}$ & $78.2_{\pm{0.8}}$ & ${69.9}_{\pm{0.9}}$ & ${69.5}_{\pm{0.9}}$ & $69.8_{\pm{1.9}}$ & $63.5_{\pm{1.1}}$ & $62.0_{\pm{0.3}}$ & ${78.9}_{\pm{6.9}}$ & $67.7_{\pm{3.9}}$ & $67.8_{\pm{4.2}}$ & ${75.1}_{\pm{4.9}}$ & $72.2_{\pm{0.1}}$ & $71.1_{\pm{0.4}}$ & ${77.1}_{\pm{3.2}}$ & $70.0_{\pm{1.3}}$ & $69.3_{\pm{1.4}}$  \\
            FedGA$\ddagger$ \cite{zhang2023fedga} & $81.7_{\pm{2.2}}$ & ${77.5}_{\pm{1.5}}$ & ${75.9}_{\pm{1.3}}$ & $77.6_{\pm{1.0}}$ & $67.7_{\pm{1.7}}$ & $66.4_{\pm{2.0}}$ & $68.8_{\pm{1.0}}$ & $64.9_{\pm{0.5}}$ & $\underline{64.7}_{\pm{1.5}}$ & $75.8_{\pm{0.9}}$ & $67.6_{\pm{1.1}}$ & $67.6_{\pm{1.4}}$ & $\underline{78.4}_{\pm{1.8}}$ & $74.4_{\pm{4.1}}$ & $74.7_{\pm{2.9}}$ & $76.6_{\pm{1.4}}$ & $70.4_{\pm{1.8}}$ & $69.9_{\pm{1.8}}$ \\
            CSAC$\ddagger$ \cite{yuan2023csac} & $83.2_{\pm{0.5}}$ & $74.9_{\pm{1.2}}$ & $75.4_{\pm{2.4}}$ & $80.8_{\pm{2.6}}$ & $71.3_{\pm{1.1}}$ & $71.5_{\pm{1.3}}$ & $67.3_{\pm{0.8}}$ & $63.9_{\pm{1.9}}$ & $61.2_{\pm{0.7}}$ & ${76.9}_{\pm{1.8}}$ & ${70.7}_{\pm{2.0}}$ & ${71.0}_{\pm{1.7}}$ & $72.4_{\pm{3.2}}$ & $69.3_{\pm{5.7}}$ & $67.5_{\pm{5.4}}$ & $76.1_{\pm{1.8}}$ & $70.0_{\pm{2.4}}$ & $69.3_{\pm{2.3}}$ \\
            FedSR \cite{nguyen2022fedsr} & $80.4_{\pm{1.1}}$ & $73.6_{\pm{2.2}}$ & $73.4_{\pm{2.1}}$ & $78.7_{\pm{2.1}}$ & $70.2_{\pm{2.2}}$ & $69.8_{\pm{2.7}}$ & $67.5_{\pm{0.6}}$ & $61.3_{\pm{1.3}}$ & $57.0_{\pm{3.8}}$ &  $80.7_{\pm{0.7}}$ & $68.0_{\pm{2.1}}$ & $68.3_{\pm{2.2}}$ & ${72.0}_{\pm{0.9}}$ & ${74.0}_{\pm{3.5}}$ & ${73.9}_{\pm{2.0}}$  & $75.9_{\pm{1.1}}$ & $69.4_{\pm{2.3}}$ & $68.5_{\pm{2.6}}$ \\
            G2G \cite{chen2024g2g} & $78.9_{\pm{1.7}}$ & $72.9_{\pm{1.9}}$ & $72.8_{\pm{2.1}}$ & $76.1_{\pm{1.8}}$ & $69.2_{\pm{0.1}}$ & $69.2_{\pm{0.1}}$ & $68.9_{\pm{2.4}}$ & $58.7_{\pm{3.5}}$ & $62.5_{\pm{1.4}}$ & $78.4_{\pm{0.3}}$ & $66.4_{\pm{2.5}}$ & $66.2_{\pm{2.2}}$ & ${76.2}_{\pm{0.2}}$ & ${72.6}_{\pm{1.4}}$ & ${71.9}_{\pm{2.1}}$  & $75.7_{\pm{1.3}}$ & $67.9_{\pm{1.9}}$ & $68.5_{\pm{1.6}}$ \\

            \hline
            RSC \cite{huang2020self} & $82.4_{\pm{1.3}}$ & $75.5_{\pm{0.9}}$ & $75.0_{\pm{1.0}}$ & $78.3_{\pm{0.6}}$ & $68.9_{\pm{1.9}}$ & $68.2_{\pm{2.6}}$ & ${70.2}_{\pm{0.8}}$ & $\underline{65.2}_{\pm{0.2}}$ & ${64.3}_{\pm{0.7}}$ & $81.8_{\pm{1.3}}$ & $71.1_{\pm{4.1}}$ & $71.4_{\pm{4.4}}$ & ${76.4}_{\pm{1.6}}$ & $\underline{77.8}_{\pm{1.1}}$ & ${76.0}_{\pm{0.5}}$ & ${77.8}_{\pm{1.1}}$ & $71.7_{\pm{1.6}}$ & $70.9_{\pm{1.8}}$ \\
            Mixstyle \cite{zhou2021domain} & $83.8_{\pm{1.5}}$ & $75.0_{\pm{1.5}}$ & $74.6_{\pm{1.7}}$ & $\underline{82.4}_{\pm{0.6}}$ & $\underline{73.4}_{\pm{0.5}}$ & $\underline{72.9}_{\pm{0.7}}$ & $68.0_{\pm{1.9}}$ & $64.0_{\pm{1.2}}$ & $62.3_{\pm{1.6}}$ & $79.0_{\pm{0.7}}$ & $\underline{74.5}_{\pm{2.3}}$ & $\underline{75.0}_{\pm{2.3}}$ & ${77.2}_{\pm{0.9}}$ & ${77.5}_{\pm{1.4}}$ & $\textbf{77.3}_{\pm{1.3}}$ & $78.1_{\pm{1.1}}$ & $\underline{72.9}_{\pm{1.4}}$ & $\underline{72.4}_{\pm{1.5}}$ \\
            ATS \cite{yang2021adversarial} & $75.5_{\pm{4.8}}$ & $73.5_{\pm{2.9}}$ & $72.2_{\pm{2.7}}$ & ${72.7}_{\pm{1.7}}$ & $70.5_{\pm{3.5}}$ & $67.9_{\pm{5.9}}$ & ${65.8}_{\pm{1.7}}$ & ${64.2}_{\pm{0.5}}$ & ${60.9}_{\pm{3.0}}$ & ${80.2}_{\pm{2.0}}$ & $70.1_{\pm{7.5}}$ & $73.4_{\pm{9.4}}$ & $75.2_{\pm{3.4}}$ & ${70.9}_{\pm{2.2}}$ & ${73.7}_{\pm{1.6}}$ & $73.9_{\pm{2.7}}$ & $69.8_{\pm{3.3}}$ & $69.6_{\pm{4.5}}$  \\
            DDAIG \cite{zhou2020deep} & $83.6_{\pm{0.0}}$ & $\underline{78.1}_{\pm{0.0}}$ & $\underline{77.3}_{\pm{0.0}}$ & $78.7_{\pm{0.2}}$ & $69.2_{\pm{2.3}}$ & $68.3_{\pm{3.1}}$ & $\textbf{71.3}_{\pm{1.1}}$ & $65.0_{\pm{0.9}}$ & $64.5_{\pm{1.3}}$ & $82.2_{\pm{1.9}}$ & $71.2_{\pm{5.5}}$ & $71.3_{\pm{5.6}}$ & $75.4_{\pm{0.8}}$ & $77.3_{\pm{1.1}}$ & $76.2_{\pm{0.5}}$ & $78.0_{\pm{0.8}}$ & $72.2_{\pm{1.9}}$ & $71.5_{\pm{2.1}}$  \\
            \hline
            \textbf{FedDAG} & $\mathbf{84.4}_{\pm{0.3}}$ & ${76.8}_{\pm{0.3}}$ & ${76.9}_{\pm{0.2}}$ & $\mathbf{82.9}_{\pm{1.4}}$ & $\mathbf{73.7}_{\pm{2.1}}$ & $\mathbf{73.6}_{\pm{2.0}}$ & $\underline{70.9}_{\pm{1.6}}$ & $\mathbf{65.8}_{\pm{0.5}}$ & ${64.0}_{\pm{1.3}}$ & $\mathbf{85.3}_{\pm{0.0}}$ & $\mathbf{78.1}_{\pm{0.0}}$ & $\mathbf{78.1}_{\pm{0.0}}$ & $\mathbf{80.7}_{\pm{1.5}}$ & $76.1_{\pm{1.2}}$ & $75.1_{\pm{0.7}}$ & $\mathbf{80.8}_{\pm{0.9}}$ & $\mathbf{74.1}_{\pm{0.8}}$ & $\mathbf{73.5}_{\pm{0.8}}$  \\
            \hline
            \bottomrule
            \end{tabular}
        }
    }
    \label{MIDOG_exp}
    \vspace{-2mm}
\end{table*}

\section{Experiment}

\subsection{Datasets, Implementation and Evaluation Metrics}

\subsubsection{Datasets} 
\begin{figure}[!tbp]
    \centering
\includegraphics[width=1\columnwidth]{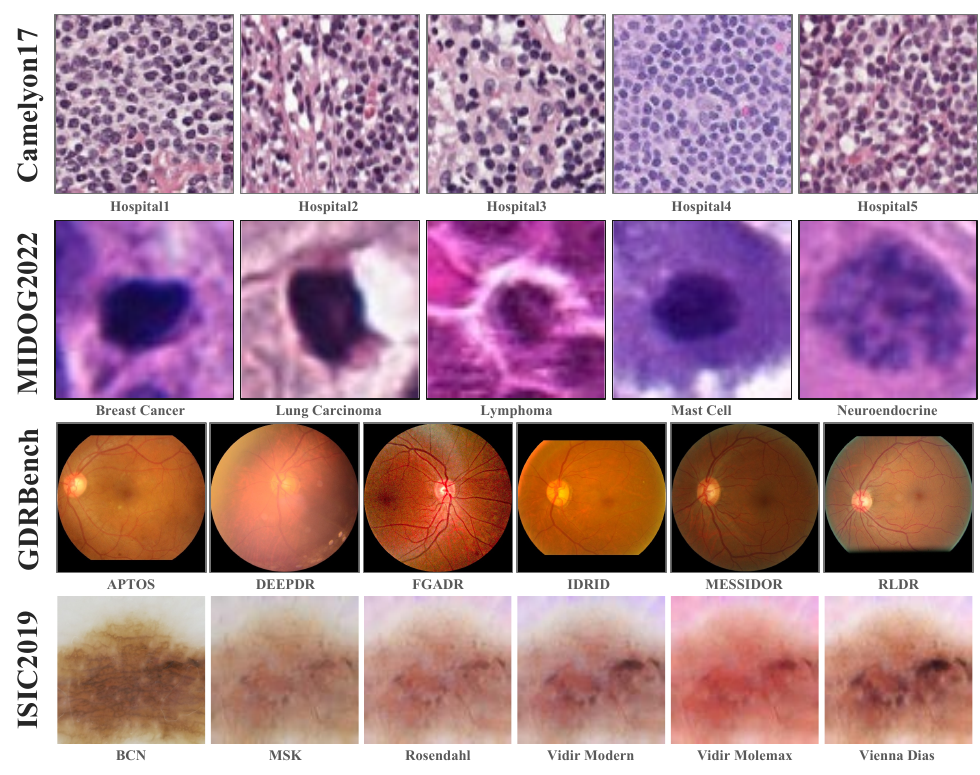} 
    \vspace{-3mm}
    \caption{{Examples from WILDS-Camelyon17, MIDOG2022, GDRBench and ISIC2019. These Benchmarks visually display subtle domain gaps.}}
    \label{fig3}
    \vspace{-3mm}
\end{figure}

{We evaluated the effectiveness of FedDAG on four renowned DG benchmarks, which contain two pathological image datasets, one fundus dataset, and one skin lesions image dataset (as illustrated in Fig. \ref{fig3}), where each benchmark covers a spectrum of medical diagnosis problems.}
\textit{1) WILDS-Camelyon17} \cite{koh2021wilds} comprises 455,954 histopathology slides from five distinct hospitals.
The labels signify whether the central region contains any tumor tissue. 
We randomly sampled ${10\%}$ of the data for our experiments.
\textit{2) MIDOG2022} \cite{aubreville2022mitosis} comprises regions of interest selected from five tumor types. 
All tumors have in common that mitotic figures are relevant for the diagnosis.
\textit{3) GDRBench} \cite{che2023dgdr} encompasses 11,357 Fundus images from six hospitals.
Labels categorize the diabetic retinopathy (DR) severity, ranging from no DR to proliferative DR.
{\textit{4) FLamby-ISIC2019} \cite{ogier2022flamby} encompasses 23,247 skin lesions images from four centers using six imaging modalities \cite{tschandl2018ham10000, codella2018skin, hernandez2024bcn20000}.}
{The labels of this dataset contain eight different types of skin lesions.}

\subsubsection{Implementation Details} 
We deployed ImageNet-pretrained ResNet50 as backbones, linear fully-connected layer as classifiers, fully convolutional networks (FCN) as generators, and adopted SGD with momentum as optimizers.
For the pathological datasets, \textit{\textit{i.e.}}, WILDS-Camelyon17 and MIDOG2022, our experiments utilized a GeForce GTX 1080Ti. 
We resized all images to 96×96 and applied normalization using ImageNet's parameters. 
Our training protocol spanned 35 communication rounds with five warmup rounds \cite{liu2021feddg,chen2023ccst}.
We employed a learning rate of 0.001, paired with a momentum of 0.9, and introduced a weight decay of $5\times10^{-4}$. 
The batch size was 32, and we standardized all loss weights to 1. 
{For the fundus and skin lesions datasets, the experiments in this section harnessed the capabilities of a GeForce RTX™ 2080Ti. 
We resized all images to 224×224 and also employed ImageNet normalization parameters. }
Our training spanned 50 communication rounds with five warmup rounds, consistent with the frameworks presented in \cite{liu2021feddg,chen2023ccst}. 
The model's learning rate was 0.005, supplemented with a momentum of 0.9. 
We also implemented a weight decay of $5\times10^{-4}$ and used a batch size 16. 
{In terms of the dataset split, we follow the original data dividing, which divides the data on each client into the training set and the validation set.}
{For datasets without a clear division of training and validation sets, such as MIDOG2022 and WILDS-Camelyon17, we randomly divide the training and validation sets in a ratio of $9:1$. }
{We determined the hyperparameters for our experiment based on the results of the model on the validation set.}
For all datasets, uniformly, all loss weights were set as 1, and our hyper-parameters, \textit{\textit{i.e.}}, $\alpha$, $\beta$, $k$, and $\rho$, were set to 0.3, 0.3, 4, and $10^{-7}$, respectively.
{To maintain fairness with other comparison methods, we selected the same hyperparameters on multiple datasets and demonstrated the effect of hyperparameter values on model performance in subsequent experiments.}

\subsubsection{Evaluation Metrics} 
For all benchmarks, we utilized a federated leave-one-domain-out evaluation protocol following \cite{liu2021feddg}. 
We reported the results of weighted accuracy (ACC), the weighted area under the ROC curve (AUC), and the weighted F1-score (F1).
The top two scores are denoted in \textbf{bold} and by \underline{underlining}.
If there is no special mention, we mainly analyze experiments via AUC results in this paper.

\subsection{Main Results}

\subsubsection{Compared Methods} 
{We compared FedDAG with extensive state-of-the-art (SOTA) personalized federated learning (PFL), DG, and FedDG methods.}
{These PFL methods include FedBN \cite{li2021fedbn}, FedProx \cite{li2020federated}, FCCL \cite{FCCL_CVPR22}, Per-FAvg \cite{fallah2020personalized}, and FedDG methods include CCST \cite{chen2023ccst}, ELCSF \cite{liu2021feddg}, FedGA \cite{zhang2023fedga}, CSAC \cite{yuan2023csac}, G2G \cite{chen2024g2g} and FedSR \cite{nguyen2022fedsr}.}
{For the DG methods such as RSC \cite{huang2020self}, Mixstyle \cite{zhou2021domain}, ATS  \cite{yang2021adversarial}, and DDAIG \cite{zhou2020deep}, we established data isolation for different clients and applied FedAvg\cite{mcmahan2017fedavg} as the communication algorithm. }

% Since AUC is more stable when evaluating complex and critical medical diagnostic challenges, we mainly analyze these methods' AUC results.

\begin{table*}[htb]
       \renewcommand\arraystretch{1.25}
        \centering
        \setlength{\tabcolsep}{2pt}
        \caption{Comparison results on the GDRBench. $\dagger$: w/ data info. sharing. $\ddagger$: w/ cross-client communication.}    
        \resizebox{1.0\textwidth}{!}{%
        \scalebox{0.8}{
        \begin{tabular}{c|ccc|ccc|ccc|ccc|ccc|ccc|ccc}
            \toprule
            % \hline
             % {} &\multicolumn{21}{C}{GDRBench}\\
            \hline
            Unseen Site &\multicolumn{3}{c}{APTOS} &\multicolumn{3}{c}{DeepDR}  &\multicolumn{3}{c}{FGADR}&\multicolumn{3}{c}{IDRID} &\multicolumn{3}{c}{Messidor}  &\multicolumn{3}{c}{RLDR} &\multicolumn{3}{c}{AVG} \\
            \hline
            Metrics  &  AUC & F1 & ACC & AUC & F1 & ACC & AUC & F1 & ACC & AUC & F1 & ACC & AUC & F1 & ACC & AUC & F1 & ACC & AUC & F1 & ACC \\
            \hline
            FedAvg \cite{mcmahan2017fedavg} & ${80.2}_{\pm{1.4}}$ & $63.3_{\pm{1.1}}$ & $59.5_{\pm{1.9}}$ & $71.7_{\pm{2.2}}$ & $56.2_{\pm{1.6}}$ & $48.6_{\pm{2.2}}$ & $64.5_{\pm{0.7}}$ & $31.7_{\pm{3.6}}$ & $28.8_{\pm{3.2}}$ & $73.8_{\pm{1.2}}$ & $45.0_{\pm{0.9}}$ & $43.5_{\pm{1.8}}$ & ${75.0}_{\pm{0.4}}$ & $64.8_{\pm{0.6}}$ & $56.8_{\pm{0.4}}$ & $55.9_{\pm{0.9}}$ & $29.6_{\pm{3.1}}$ & $21.7_{\pm{3.1}}$ & $70.2_{\pm{1.1}}$ & $48.4_{\pm{1.8}}$ & $43.1_{\pm{2.1}}$ \\
            \hline
            FedBN \cite{li2021fedbn} & $80.3_{\pm{0.3}}$ & $63.0_{\pm{2.0}}$ & $60.0_{\pm{3.2}}$ & $75.0_{\pm{ 2.9}}$ & $56.3_{\pm{2.1}}$ & $48.7_{\pm{2.5}}$ & $66.8_{\pm{0.8}}$ & $33.8_{\pm{3.5}}$ & ${31.9}_{\pm{2.8}}$ & $77.0_{\pm{1.1}}$ & $42.4_{\pm{3.3}}$ & $41.6_{\pm{3.3}}$ & ${77.9}_{\pm{0.6}}$ & $64.7_{\pm{0.7}}$ & $57.3_{\pm{0.2}}$ & $58.5_{\pm{1.7}}$ & $29.6_{\pm{2.3}}$ & ${31.7}_{\pm{1.2}}$ & $72.6_{\pm{1.2}}$ & $48.5_{\pm{2.3}}$ & $45.1_{\pm{2.2}}$ \\
            FedProx \cite{li2020federated} & $\underline{82.5}_{\pm{1.4}}$ & $63.1_{\pm{1.5}}$ & $59.4_{\pm{2.8}}$ & $73.9_{\pm{2.4}}$ & $55.7_{\pm{1.0}}$ & $48.5_{\pm{1.1}}$ & $67.8_{\pm{0.3}}$ & ${34.2}_{\pm{1.4}}$ & $31.0_{\pm{0.9}}$ & $77.1_{\pm{1.2}}$ & $47.6_{\pm{3.2}}$ & $45.0_{\pm{1.8}}$ & $78.3_{\pm{0.7}}$ & $64.6_{\pm{0.3}}$ & $57.2_{\pm{0.8}}$ & $59.2_{\pm{1.1}}$ & $27.6_{\pm{1.7}}$ & $24.7_{\pm{1.1}}$ & $73.1_{\pm{1.2}}$ & ${48.8}_{\pm{1.5}}$ & $44.3_{\pm{1.4}}$ \\
            FCCL \cite{FCCL_CVPR22}  & $76.1_{\pm{2.6}}$ & $60.0_{\pm{1.8}}$ & $51.8_{\pm{2.4}}$ & $69.3_{\pm{6.3}}$ & $51.2_{\pm{2.5}}$ & $39.8_{\pm{3.3}}$ & $66.5_{\pm{17}}$ & ${35.1}_{\pm{1.1}}$ & ${45.5}_{\pm{0.5}}$ & $69.8_{\pm{3.2}}$ & $36.6_{\pm{5.0}}$ & ${42.1}_{\pm{7.9}}$ & $68.6_{\pm{1.4}}$ & $60.0_{\pm{1.3}}$ & $50.7_{\pm{0.3}}$ & $56.9_{\pm{3.0}}$ & $20.9_{\pm{16.0}}$ & $15.1_{\pm{19.0}}$ & $67.9_{\pm{3.0}}$ & $44.1_{\pm{4.6}}$ & $40.8_{\pm{5.6}}$ \\
            Per-FAvg \cite{fallah2020personalized}  & $78.8_{\pm{1.8}}$ & $50.8_{\pm{0.7}}$ & $55.4_{\pm{7.1}}$ & $73.5_{\pm{2.4}}$ & $50.8_{\pm{0.7}}$ & $42.0_{\pm{2.7}}$ & $70.1_{\pm{7.0}}$ & $ \underline{44.1}_{\pm{10.1}}$ & $42.0_{\pm{2.7}}$ & $63.0_{\pm{5.9}}$ & $34.0_{\pm{3.6}}$ & $20.4_{\pm{7.1}}$ & $69.0_{\pm{3.0}}$ & $54.0_{\pm{7.9}}$ & $47.8_{\pm{3.8}}$ & $52.8_{\pm{5.2}}$ & ${23.1}_{\pm{18.3}}$ & $16.0_{\pm{18.8}}$ & $67.9_{\pm{4.2}}$ & $42.8_{\pm{7.2}}$ & $37.4_{\pm{7.1}}$ \\
            \hline
            CCST$\dagger$ \cite{chen2023ccst} & $80.7_{\pm{1.0}}$ & $62.1_{\pm{0.9}}$ & $58.2_{\pm{1.9}}$ & $\underline{77.5}_{\pm{1.1}}$ & $49.2_{\pm{0.0}}$ & $52.2_{\pm{0.4}}$ & $68.0_{\pm{0.8}}$ & $31.3_{\pm{1.7}}$ & $37.7_{\pm{1.6}}$ & $76.3_{\pm{0.3}}$ & $\textbf{49.5}_{\pm{0.8}}$ & $40.6_{\pm{2.7}}$ & $76.4_{\pm{0.5}}$ & $54.8_{\pm{1.2}}$ & $57.9_{\pm{1.2}}$ & $64.8_{\pm{0.8}}$ & $31.4_{\pm{2.9}}$ & $24.4_{\pm{3.0}}$ & $73.9_{\pm{0.8}}$ & $46.3_{\pm{1.3}}$ & $45.2_{\pm{1.8}}$ \\
            ELCFS$\dagger$ \cite{liu2021feddg} & $76.6_{\pm{2.1}}$ & ${61.1}_{\pm{4.1}}$ & ${54.1}_{\pm{6.3}}$ & $69.7_{\pm{2.3}}$ & $55.6_{\pm{0.7}}$ & $47.1_{\pm{0.2}}$ & ${69.6}_{\pm{0.9}}$ & $\textbf{59.3}_{\pm{2.9}}$ & $\textbf{49.1}_{\pm{0.7}}$ & $72.7_{\pm{1.1}}$ & $34.7_{\pm{2.1}}$ & $35.4_{\pm{2.9}}$ & ${69.1}_{\pm{0.5}}$ & $58.0_{\pm{2.8}}$ & $48.9_{\pm{0.6}}$ & $63.0_{\pm{0.9}}$ & $23.8_{\pm{2.4}}$ & $24.5_{\pm{3.3}}$ & ${70.3}_{\pm{1.3}}$ & $ 48.8_{\pm{2.5}} $ & $ 43.2_{\pm{2.3}}$ \\
            FedGA$\ddagger$ \cite{zhang2023fedga} & $77.5_{\pm{1.3}}$ & $54.5_{\pm{4.6}}$ & $54.4_{\pm{4.1}}$ & $73.8_{\pm{1.5}}$ & $55.2_{\pm{2.4}}$ & $50.5_{\pm{2.9}}$ & $68.5_{\pm{2.0}}$ & $39.4_{\pm{2.3}}$ & $37.8_{\pm{3.3}}$ & $77.5_{\pm{1.3}}$ & $25.6_{\pm{3.3}}$ & $31.1_{\pm{1.7}}$ & $71.7_{\pm{0.8}}$ & $54.7_{\pm{2.7}}$ & $57.1_{\pm{3.5}}$ & $61.8_{\pm{3.0}}$ & $19.5_{\pm{2.3}}$ & $14.4_{\pm{1.7}}$ & $71.8_{\pm{1.7}}$ & $41.5_{\pm{2.9}}$ & $40.8_{\pm{2.7}}$ \\
            CSAC$\ddagger$ \cite{yuan2023csac}  & $69.7_{\pm{2.4}}$ & $37.8_{\pm{4.8}}$ & $43.5_{\pm{4.5}}$ & $72.0_{\pm{2.0}}$ & $37.3_{\pm{1.2}}$ & $45.1_{\pm{3.5}}$ & $62.2_{\pm{1.9}}$ & $34.2_{\pm{1.3}}$ & $\underline{46.0}_{\pm{0.6}}$ & ${71.1}_{\pm{3.2}}$ & $47.1_{\pm{2.1}}$ & $\textbf{47.4}_{\pm{4.6}}$ & $70.3_{\pm{0.5}}$ & $45.2_{\pm{5.3}}$ & $52.1_{\pm{5.0}}$ & ${64.8}_{\pm{0.6}}$ & $\underline{23.8}_{\pm{4.6}}$ & ${31.7}_{\pm{1.9}}$ & $68.4_{\pm{1.8}}$ & $37.6_{\pm{3.2}}$ & $44.3_{\pm{3.4}}$ \\
            FedSR \cite{nguyen2022fedsr} & $62.9_{\pm{0.9}}$ & $39.8_{\pm{9.1}}$ & $41.9_{\pm{6.3}}$ & $53.5_{\pm{2.4}}$ & $31.5_{\pm{0.6}}$ & $23.6_{\pm{3.2}}$ & $51.7_{\pm{1.5}}$ & $39.6_{\pm{1.6}}$ & $37.9_{\pm{0.4}}$ & $66.5_{\pm{1.6}}$ & $\underline{47.6}_{\pm{2.1}}$ & $41.8_{\pm{5.3}}$ & $60.7_{\pm{2.8}}$ & $34.2_{\pm{1.0}}$ & $\underline{59.0}_{\pm{2.3}}$ & $\textbf{72.4}_{\pm{6.3}}$ & $\textbf{47.5}_{\pm{17.7}}$ & $\textbf{43.3}_{\pm{14.0}}$ & $61.3_{\pm{2.6}}$ & $40.0_{\pm{5.4}}$ & $41.3_{\pm{5.3}}$ \\
            G2G \cite{chen2024g2g} & $74.2_{\pm{1.5}}$ & $48.8_{\pm{4.3}}$ & $47.6_{\pm{5.1}}$ & $70.8_{\pm{1.0}}$ & $42.9_{\pm{0.5}}$ & $51.2_{\pm{0.8}}$ & $65.9_{\pm{1.2}}$ & $28.2_{\pm{2.4}}$ & $28.3_{\pm{2.2}}$ & $77.2_{\pm{1.3}}$ & ${44.8}_{\pm{4.6}}$ & $\underline{46.1}_{\pm{0.3}}$ & $\textbf{79.3}_{\pm{0.5}}$ & $55.4_{\pm{0.5}}$ & $\textbf{63.7}_{\pm{0.6}}$ & ${60.4}_{\pm{0.3}}$ & ${17.8}_{\pm{2.4}}$ & ${19.6}_{\pm{1.3}}$ & $71.3{\pm{1.0}}$ & $39.7_{\pm{2.4}}$ & $42.8_{\pm{2.7}}$ \\

            \hline

            RSC \cite{huang2020self} & $80.8_{\pm{1.0}}$ & $62.0_{\pm{2.1}}$ & $58.3_{\pm{1.8}}$ & $75.5_{\pm{0.3}}$ & $\textbf{59.8}_{\pm{1.2}}$ & $\underline{52.6}_{\pm{1.7}}$ & $67.0_{\pm{1.0}}$ & $38.1_{\pm{2.6}}$ & $34.6_{\pm{1.4}}$ & $76.5_{\pm{1.6}}$ & $47.6_{\pm{2.1}}$ & $41.8_{\pm{5.3}}$ & $76.4_{\pm{0.9}}$ & ${64.8}_{\pm{0.9}}$ & ${55.9}_{\pm{1.1}}$ & ${62.9}_{\pm{0.9}}$ & $25.5_{\pm{1.1}}$ & $26.8_{\pm{1.3}}$ & $72.2_{\pm{1.0}}$ & $49.6_{\pm{1.7}}$ & $45.0_{\pm{2.1}}$ \\
            Mixstyle \cite{zhou2021domain}  & $81.9_{\pm{0.4}}$ & $\underline{63.8}_{\pm{1.2}}$ & $\underline{61.4}_{\pm{1.2}}$ & $76.0_{\pm{0.5}}$ & $56.2_{\pm{1.3}}$ & $51.0_{\pm{1.2}}$ & $71.3_{\pm{1.3}}$ & ${37.8}_{\pm{1.6}}$ & $34.8_{\pm{2.2}}$ & $75.1_{\pm{1.1}}$ & ${45.6}_{\pm{1.6}}$  & $36.8_{\pm{0.7}}$ & $76.4_{\pm{1.3}}$ & $\textbf{66.5}_{\pm{0.7}}$ & ${58.5}_{\pm{0.6}}$ & $65.2_{\pm{0.2}}$ & $30.0_{\pm{4.2}}$ & $32.3_{\pm{5.2}}$ & $\underline{74.3}_{\pm{0.8}}$ & $\underline{49.9}_{\pm{1.8}}$& $45.8_{\pm{1.9}}$ \\
            
            ATS \cite{yang2021adversarial}  & ${78.7}_{\pm{4.3}}$ & $61.9_{\pm{6.8}}$ & $59.7_{\pm{6.5}}$ & $72.1_{\pm{1.2}}$ & $47.7_{\pm{2.5}}$ & $49.7_{\pm{5.0}}$ & $\textbf{74.0}_{\pm{0.2}}$ & $43.3_{\pm{0.6}}$ & $43.1_{\pm{0.3}}$ & $72.4_{\pm{0.7}}$ & $45.5_{\pm{1.3}}$ & ${45.5}_{\pm{2.4}}$ & $72.9_{\pm{0.7}}$ & $51.1_{\pm{2.2}}$ & $52.4_{\pm{1.1}}$ & $60.4_{\pm{0.6}}$ & $28.4_{\pm{2.1}}$ & ${30.4}_{\pm{2.6}}$ & ${72.8}_{\pm{1.3}}$ & $46.2_{\pm{2.6}}$ & $\underline{46.8}_{\pm{3.0}}$ \\
            DDAIG \cite{zhou2020deep} & $80.3_{\pm{1.3}}$ & $61.0_{\pm{2.0}}$ & $56.0_{\pm{2.1}}$ & ${77.3}_{\pm{0.8}}$ & $\underline{58.2}_{\pm{1.3}}$ & ${50.7}_{\pm{1.7}}$ & $68.3_{\pm{0.5}}$ & $39.3_{\pm{2.2}}$ & $38.0_{\pm{2.8}}$ & $\underline{77.6}_{\pm{0.6}}$ & $42.8_{\pm{7.8}}$ & $38.0_{\pm{4.2}}$ & $75.4_{\pm{0.7}}$ & $62.5_{\pm{1.3}}$ & $53.1_{\pm{1.4}}$ & $54.8_{\pm{3.2}}$ & $28.2_{\pm{1.5}}$ & $27.5_{\pm{3.2}}$ & $72.3_{\pm{1.2}}$ & ${48.7}_{\pm{2.7}}$ & ${43.9}_{\pm{2.5}}$ \\
            \hline
            \textbf{FedDAG} & $\textbf{83.6}_{\pm{1.4}}$ & $\textbf{65.3}_{\pm{1.1}}$ & $\textbf{65.5}_{\pm{1.9}}$ & $\textbf{82.8}_{\pm{2.0}}$ & ${56.2}_{\pm{1.6}}$ & $\textbf{58.6}_{\pm{2.2}}$ & $\underline{73.2}_{\pm{1.5}}$ & $36.9_{\pm{1.1}}$ & ${40.7}_{\pm{1.6}}$ & $\textbf{81.2}_{\pm{1.6}}$ & $45.0_{\pm{0.9}}$ & $43.5_{\pm{1.8}}$ & $\underline{78.0}_{\pm{0.4}}$ & $\underline{64.8}_{\pm{0.6}}$ & ${56.8}_{\pm{0.4}}$ & $\underline{70.7}_{\pm{0.7}}$ & $\underline{33.7}_{\pm{3.3}}$ & $\underline{33.7}_{\pm{2.8}}$ & $\textbf{78.3}_{\pm{1.3}}$ & $\textbf{50.4}_{\pm{1.4}}$ & $\textbf{49.8}_{\pm{1.8}}$ \\
            \hline
            \bottomrule
            \end{tabular}
            \label{table Supplement}
    }}
    \label{GDR_exp}
    \vspace{-3mm}
\end{table*}

\begin{table*}[htb]
    \caption{Comparison results on the FLamby-ISIC2019. $\dagger$: w/ data info. sharing. $\ddagger$: w/ cross-client communication.}
    % \vspace{-2mm}
    \renewcommand\arraystretch{1.25}
    \centering
    \setlength{\tabcolsep}{2pt}
    \resizebox{1.0\textwidth}{!}{
        \scalebox{0.8}{
            \begin{tabular}{c|ccc|ccc|ccc|ccc|ccc|ccc|ccc} 
            \toprule
            % \hline
            % {}&\multicolumn{18}{c}{WILDS-Camelyon17}\\
            \hline
            Unseen Site &\multicolumn{3}{c}{BCN}&\multicolumn{3}{c}{HAM Vidir Molemax}&\multicolumn{3}{c}{HAM Vidir Modern}&\multicolumn{3}{c}{HAM Rosendahl}&\multicolumn{3}{c}{MSK}&\multicolumn{3}{c}{HAM Vienna Dias}&\multicolumn{3}{c}{AVG} \\
            \hline      
            Metrics  & AUC & F1 & ACC & AUC & F1 & ACC & AUC & F1 & ACC & AUC & F1 & ACC & AUC & F1 & ACC & AUC & F1 & ACC & AUC & F1 & ACC \\
            \hline
            FedAvg \cite{mcmahan2017fedavg} & $70.5_{\pm{0.2}}$ & $34.9_{\pm{0.6}}$ & $40.6_{\pm{0.6}}$& $\underline{95.6}_{\pm{0.4}}$ & $94.3_{\pm{0.0}}$ & $92.6_{\pm{0.8}}$ & $\underline{90.7}_{\pm{0.9}}$ & $\underline{70.1}_{\pm{0.2}}$ & ${72.5}_{\pm{0.4}}$ & $73.7_{\pm{0.4}}$ & $25.9_{\pm{0.3}}$ & $40.2_{\pm{0.8}}$ & $73.6_{\pm{0.8}}$ & $12.1_{\pm{0.0}}$ & $27.8_{\pm{0.1}}$ & $95.4_{\pm{0.2}}$ & $\underline{78.8}_{\pm{0.1}}$ & ${76.9}_{\pm{0.5}}$ & $83.3_{\pm{0.2}}$ & $52.7_{\pm{0.1}}$ & $58.4_{\pm{0.5}}$\\
            % \hline
            FedBN \cite{li2021fedbn} & $74.2_{\pm{0.2}}$ & $36.5_{\pm{0.1}}$ & $42.1_{\pm{0.2}}$ & $95.4_{\pm{0.5}}$ & $\textbf{95.8}_{\pm{1.7}}$ & $\textbf{95.9}_{\pm{2.3}}$ & $87.0_{\pm{0.4}}$ & $61.1_{\pm{1.1}}$ & $67.1_{\pm{1.2}}$ & $77.5_{\pm{0.6}}$ & $34.1_{\pm{0.3}}$ & $45.1_{\pm{0.4}}$ & $\underline{90.5}_{\pm{0.7}}$ & $36.6_{\pm{2.1}}$ & $44.8_{\pm{2.6}}$ & $\underline{95.9}_{\pm{0.5}}$ & $77.6_{\pm{1.1}}$ & $76.2_{\pm{1.3}}$ & $86.7_{\pm{0.5}}$ & $57.1_{\pm{1.1}}$ & $61.9_{\pm{1.3}}$  \\
            FedProx \cite{li2020federated} & $65.8_{\pm{0.2}}$ & $16.5_{\pm{2.5}}$ & $33.2_{\pm{2.5}}$ & $86.1_{\pm{0.5}}$ & $90.7_{\pm{2.1}}$ & $93.6_{\pm{2.3}}$ & $73.8_{\pm{1.6}}$ & $41.9_{\pm{3.0}}$ & $57.4_{\pm{3.2}}$ & $69.6_{\pm{0.6}}$ & $21.0_{\pm{1.7}}$ & $38.0_{\pm{1.8}}$ & $70.8_{\pm{1.5}}$ & $12.1_{\pm{2.3}}$ & $27.8_{\pm{1.9}}$ & $69.2_{\pm{0.9}}$ & $49.3_{\pm{1.0}}$ & $63.5_{\pm{1.1}}$  & $72.6_{\pm{0.9}}$ & $38.6_{\pm{1.2}}$ & $52.3_{\pm{2.1}}$ \\
            FCCL \cite{FCCL_CVPR22} & $66.0_{\pm{2.7}}$ & $17.0_{\pm{0.6}}$ & $33.4_{\pm{2.9}}$ & $87.4_{\pm{1.6}}$ & $90.7_{\pm{0.2}}$ & $93.6_{\pm{0.5}}$ & $72.4_{\pm{1.1}}$ & $41.9_{\pm{0.3}}$ & $57.4_{\pm{0.4}}$ & $68.5_{\pm{1.5}}$ & $21.0_{\pm{1.8}}$ & $38.0_{\pm{3.1}}$ & $67.3_{\pm{2.9}}$ & $13.0_{\pm{3.1}}$ & $28.9_{\pm{3.1}}$ & $75.2_{\pm{2.1}}$ & $52.6_{\pm{1.1}}$ & $65.4_{\pm{1.8}}$ & $73.5_{\pm{2.0}}$ & $39.4_{\pm{1.2}}$ & $52.3_{\pm{1.8}}$\\
            Per-FAvg \cite{fallah2020personalized} & $65.7_{\pm{0.2}}$ & $16.5_{\pm{0.5}}$ & $33.2_{\pm{1.2}}$ & $84.7_{\pm{0.1}}$ & $91.2_{\pm{1.1}}$ & $94.6_{\pm{0.5}}$ & $72.1_{\pm{0.2}}$ & $42.5_{\pm{0.6}}$ & $57.7_{\pm{1.1}}$ & $69.5_{\pm{1.1}}$ & $21.0_{\pm{1.7}}$ & $38.1_{\pm{2.3}}$ & $70.8_{\pm{2.0}}$ & $12.1_{\pm{1.6}}$ & $27.8_{\pm{0.7}}$ & $74.5_{\pm{0.5}}$ & $49.3_{\pm{1.1}}$ & $63.5_{\pm{0.7}}$  & $72.9_{\pm{0.5}}$ & $38.8_{\pm{1.1}}$ & $52.5_{\pm{0.7}}$\\
            \hline
            CCST$\dagger$ \cite{chen2023ccst} & $\underline{75.0}_{\pm{0.3}}$ & $37.7_{\pm{0.5}}$ & $42.3_{\pm{0.5}}$ & ${95.4}_{\pm{0.2}}$ & ${94.2}_{\pm{2.2}}$ & ${93.5}_{\pm{1.9}}$ & ${89.1}_{\pm{0.4}}$ & ${51.4}_{\pm{2.5}}$ & $\textbf{73.3}_{\pm{2.3}}$ & $75.6_{\pm{0.1}}$ & ${44.5}_{\pm{2.5}}$ & ${48.5}_{\pm{2.3}}$ & $73.9_{\pm{0.8}}$ & $\textbf{50.9}_{\pm{2.1}}$ & $\textbf{54.4}_{\pm{2.2}}$ & $88.6_{\pm{0.4}}$ & ${70.1}_{\pm{1.9}}$ & ${72.7}_{\pm{1.8}}$ & $82.9_{\pm{0.3}}$ & ${58.1}_{\pm{2.0}}$ & $\underline{64.1}_{\pm{1.7}}$ \\
            ELCFS$\dagger$ \cite{liu2021feddg} & ${70.8}_{\pm{0.8}}$ & ${20.5}_{\pm{0.9}}$ & ${34.5}_{\pm{1.0}}$ & $93.1_{\pm{0.7}}$ & $\underline{94.3}_{\pm{0.9}}$ & $\underline{95.7}_{\pm{1.0}}$ & $83.8_{\pm{1.5}}$ & $49.9_{\pm{2.1}}$ & $61.1_{\pm{2.4}}$ & ${75.1}_{\pm{0.8}}$ & $31.2_{\pm{4.1}}$ & $42.7_{\pm{4.6}}$ & ${74.8}_{\pm{0.9}}$ & $17.3_{\pm{1.4}}$ & $31.1_{\pm{1.6}}$ & ${87.1}_{\pm{0.9}}$ & $60.1_{\pm{1.9}}$ & $67.3_{\pm{2.1}}$ & ${80.8}_{\pm{0.9}}$ & $45.5_{\pm{1.8}}$ & $55.4_{\pm{2.2}}$    \\
            FedGA$\ddagger$ \cite{zhang2023fedga} & $73.9_{\pm{0.1}}$ & $\textbf{41.9}_{\pm{2.5}}$ & $\underline{45.9}_{\pm{3.0}}$ & $94.1_{\pm{0.1}}$ & $93.2_{\pm{2.6}}$ & $91.6_{\pm{3.0}}$ & $87.2_{\pm{0.1}}$ & $67.3_{\pm{3.3}}$ & $68.7_{\pm{3.8}}$ & ${80.1}_{\pm{0.5}}$ & $\textbf{51.2}_{\pm{3.2}}$ & ${53.4}_{\pm{3.8}}$ & $86.7_{\pm{1.7}}$ & $35.3_{\pm{2.4}}$ & $42.2_{\pm{2.7}}$ & $94.9_{\pm{0.9}}$ & $72.7_{\pm{2.8}}$ & $71.1_{\pm{3.3}}$ & $86.1_{\pm{0.9}}$ & $60.3_{\pm{2.8}}$ & $62.2_{\pm{3.3}}$ \\
            CSAC$\ddagger$ \cite{yuan2023csac} & $64.7_{\pm{2.1}}$ & ${38.4}_{\pm{4.1}}$ & ${44.1}_{\pm{6.3}}$ & $84.2_{\pm{2.3}}$ & $94.6_{\pm{0.7}}$ & $95.5_{\pm{0.2}}$ & ${79.5}_{\pm{0.9}}$ & ${66.4}_{\pm{2.9}}$ & ${69.3}_{\pm{0.7}}$ & $69.3_{\pm{1.1}}$ & $48.3_{\pm{2.1}}$ & $45.9_{\pm{2.9}}$ & ${80.1}_{\pm{0.5}}$ & $\underline{41.2}_{\pm{2.8}}$ & $47.8_{\pm{0.6}}$ & $83.8_{\pm{0.9}}$ & $76.4_{\pm{2.4}}$ & $73.1_{\pm{3.3}}$ & ${76.9}_{\pm{1.3}}$ & $60.0_{\pm{2.5}}$ & $62.6_{\pm{2.3}}$ \\
            FedSR \cite{nguyen2022fedsr} & $74.1_{\pm{1.8}}$ & $\underline{41.7}_{\pm{0.7}}$ & $45.9_{\pm{7.1}}$ & $95.4_{\pm{2.4}}$ & $95.8_{\pm{0.7}}$ & $95.7_{\pm{2.7}}$ & $87.1_{\pm{7.0}}$ & ${67.7}_{\pm{10.1}}$ & $69.9_{\pm{2.7}}$ & $80.5_{\pm{5.9}}$ & $50.2_{\pm{3.6}}$ & $\underline{54.3}_{\pm{7.1}}$ & $82.5_{\pm{3.0}}$ & $32.0_{\pm{7.9}}$ & $42.2_{\pm{3.8}}$ & $94.8_{\pm{5.2}}$ & ${78.6}_{\pm{6.3}}$ & $76.9_{\pm{6.8}}$ & $85.7_{\pm{4.2}}$ & $60.0_{\pm{5.2}}$ & $64.0_{\pm{5.1}}$  \\
            G2G \cite{chen2024g2g} & $71.8_{\pm{0.6}}$ & ${32.7}_{\pm{1.1}}$ & $40.3_{\pm{0.7}}$ & $94.4_{\pm{0.1}}$ & $94.9_{\pm{0.0}}$ & $95.1_{\pm{0.0}}$ & $88.6_{\pm{0.3}}$ & ${65.7}_{\pm{0.8}}$ & $68.2_{\pm{0.5}}$ & $68.6_{\pm{0.7}}$ & $22.1_{\pm{0.3}}$ & ${38.1}_{\pm{0.0}}$ & $60.1_{\pm{4.8}}$ & $14.5_{\pm{1.7}}$  & $25.4_{\pm{1.7}}$ & $69.5_{\pm{0.8}}$ & $50.8_{\pm{2.2}}$ & ${57.7}_{\pm{1.3}}$ & $75.5_{\pm{1.2}}$ & $46.8_{\pm{1.0}}$ & $54.1_{\pm{0.7}}$  \\
            \hline
            ATS \cite{yang2021adversarial} & $62.6_{\pm{1.2}}$ & $20.7_{\pm{1.1}}$ & $34.5_{\pm{1.2}}$ & ${94.5}_{\pm{2.2}}$ & $88.1_{\pm{1.9}}$ & $88.9_{\pm{1.5}}$ & ${81.3}_{\pm{0.1}}$ & ${49.7}_{\pm{0.2}}$ & ${59.4}_{\pm{0.2}}$ & ${76.8}_{\pm{0.4}}$ & $40.4_{\pm{0.9}}$ & $44.4_{\pm{1.2}}$ & $79.4_{\pm{1.7}}$ & ${27.9}_{\pm{0.6}}$ & ${35.6}_{\pm{1.7}}$ & $86.1_{\pm{1.1}}$ & $59.5_{\pm{2.1}}$ & $67.3_{\pm{1.2}}$ & $79.6_{\pm{1.1}}$ & $47.7_{\pm{2.1}}$ & $55.0_{\pm{1.2}}$  \\
            RSC \cite{huang2020self} & $64.8_{\pm{1.0}}$ & $16.8_{\pm{2.1}}$ & $33.3_{\pm{1.8}}$ & $87.8_{\pm{0.3}}$ & ${91.3}_{\pm{1.2}}$ & ${93.8}_{\pm{1.7}}$ & $73.6_{\pm{1.0}}$ & $42.4_{\pm{2.6}}$ & $57.3_{\pm{1.4}}$ & $71.3_{\pm{1.6}}$ & $24.8_{\pm{2.1}}$ & $40.2_{\pm{5.3}}$ & $69.6_{\pm{0.9}}$ & ${12.1}_{\pm{0.9}}$ & ${27.8}_{\pm{1.1}}$ & ${77.4}_{\pm{0.9}}$ & $50.5_{\pm{1.1}}$ & $63.5_{\pm{1.3}}$ & $74.1_{\pm{1.0}}$ & $39.7_{\pm{1.7}}$ & $52.7_{\pm{2.1}}$ \\
            Mixstyle \cite{zhou2021domain}  & $66.0_{\pm{0.4}}$ & ${17.0}_{\pm{1.2}}$ & ${33.4}_{\pm{1.2}}$ & $87.4_{\pm{0.5}}$ & $90.7_{\pm{1.3}}$ & $93.6_{\pm{1.2}}$ & $72.4_{\pm{1.3}}$ & ${41.9}_{\pm{1.6}}$ & $57.4_{\pm{2.2}}$ & $67.3_{\pm{1.1}}$ & ${21.0}_{\pm{1.6}}$  & $38.0_{\pm{0.7}}$ & $69.3_{\pm{1.3}}$ & ${12.1}_{\pm{0.7}}$ & ${28.9}_{\pm{0.6}}$ & $74.5_{\pm{0.2}}$ & $49.3_{\pm{4.2}}$ & $63.5_{\pm{5.2}}$ & ${72.8}_{\pm{0.8}}$ & ${38.7}_{\pm{1.8}}$ & $52.5_{\pm{1.9}}$ \\
            DDAIG \cite{zhou2020deep} & $72.7_{\pm{0.4}}$ & $36.7_{\pm{0.9}}$ & $41.7_{\pm{1.1}}$ & $95.0_{\pm{0.2}}$ & $93.7_{\pm{3.1}}$ & $92.3_{\pm{3.5}}$ & $89.4_{\pm{0.6}}$ & $66.5_{\pm{3.8}}$ & $69.3_{\pm{4.3}}$ & $\underline{81.6}_{\pm{0.3}}$ & $\underline{51.1}_{\pm{2.8}}$ & $\textbf{55.6}_{\pm{3.1}}$ & $90.1_{\pm{1.3}}$ & ${41.1}_{\pm{1.8}}$ & $\underline{47.8}_{\pm{1.9}}$ & $91.4_{\pm{0.6}}$ & $75.8_{\pm{2.5}}$ & $\underline{76.9}_{\pm{2.8}}$ & $\underline{86.7}_{\pm{0.4}}$ & $\underline{60.3}_{\pm{2.2}}$ & $63.9_{\pm{2.5}}$  \\
            \hline
            \textbf{FedDAG} & $\textbf{76.6}_{\pm{0.5}}$ & ${40.1}_{\pm{0.4}}$ & $\textbf{46.9}_{\pm{0.4}}$ & $\textbf{95.6}_{\pm{0.2}}$ & ${92.8}_{\pm{0.7}}$ & ${87.2}_{\pm{0.7}}$ & $\textbf{92.3}_{\pm{0.7}}$ & $\textbf{71.4}_{\pm{3.1}}$ & $\underline{72.9}_{\pm{3.3}}$ & $\textbf{81.6}_{\pm{0.4}}$ & $43.2_{\pm{0.6}}$ & ${53.3}_{\pm{0.7}}$ & $\textbf{91.4}_{\pm{0.5}}$ & $33.5_{\pm{2.1}}$ & ${46.2}_{\pm{2.3}}$ & $\textbf{95.9}_{\pm{0.1}}$ & $\textbf{81.8}_{\pm{0.8}}$ & $\textbf{84.8}_{\pm{0.9}}$ & $\textbf{89.1}_{\pm{0.4}}$ & $\textbf{60.5}_{\pm{1.3}}$ & $\textbf{65.3}_{\pm{1.4}}$  \\
            \hline
            \bottomrule
            \end{tabular}
        }
    }
    \label{ISIC_exp}
    \vspace{-2mm}
\end{table*}

\subsubsection{Evaluation on Pathological Benchmarks} 
{Table \ref{Camelyon_exp} exhibits the comparison results of FedDAG against other SOTA methods on WILDS-Camelyon17.}
{Due to the data isolation in FL scenarios, DG methods may be sub-optimal for generalization capability under the federated setting.}
This view is proved by the results of ATS and DDAIG, which perform on FL scenarios and merely bring ${2.8\%}$ and ${2.9\%}$ performance increases, respectively.
Simultaneously, we can observe the limited improvement of PFL approaches, such as FCCL and Per-FAvg.
{This may be because PFL focuses on enhancing the generalization to source clients, which neglects the generalization of the target client.}
As for the FedDG approaches, FedGA enhances ${3.1\%}$ relative to baseline, where the advancement can be attributed to its emphasis on mitigating distinct contributions among clients \cite{zhang2023fedga}.
{CCST and ELCFS focus on increasing data diversity and simulating potential domain gaps \cite{liu2021feddg, chen2023ccst}, which effectively and stably improved the generalization ability of the model.}
They brought performance improvements of ${3.7\%}$ and ${4.0\%}$, respectively, proving the effectiveness of recombining domain attributes. 
{However, these methods are sub-optimal in generalizing to the unseen target domain because they do not touch the out-of-distribution of global data distribution.}
FedDAG, while building upon these insights, surpasses these methods markedly, which achieves a performance improvement of ${5.3\%}$.
{The experimental results of MIDOG2022 are displayed in Table  \ref{MIDOG_exp}.}
We can observe results and conduct conclusions similar to those on WILDS-Camelyon17.
{However, due to more ambiguous domain gaps than WILDS-Camelyon17, there are some different trends.}
{For example, FedGA only brings ${0.1\%}$ performance improvement, which may be due to the local model biases towards the local data distribution, resulting in limited improvement in the generalization ability of the global model.}
The limited performance improvement of ELCFS also illustrates that recombining domain attributes is insufficient to approximate the unseen domain shifts when ambiguous domain gaps exist.
{However, despite the ambiguous domain gap, FedDAG still performs well and shows $4.3\%$ improvement over the baseline.}
The experimental results on two pathology datasets illustrate that FedDAG can significantly improve performance regardless of the presence or absence of ambiguous domain gaps.
{This superior performance improvement of FedDAG is credited to the holistic design of mutual boost between NDAG and SHA.}

\subsubsection{Evaluation on GDRBench} 
{The comparison results of FedDAG and other SOTA methods on GDRBench are shown}
{in Table \ref{GDR_exp}.}
{We can observe some similar conclusions to pathological benchmarks, and there are also some different conclusions.}
For example, ELCFS is an extremely limited improvement compared to the baseline.
This may be since recombining frequency domain information only expands the local data without touching the out-of-distribution of global data.
{Consequently, it fails to approximate unseen domain shifts.
The methods of novel domain generation merely within the local data distribution and perform well, \textit{i.e.}, ATS and DDAIG, were enhanced by ${2.6\%}$ and ${2.1\%}$ in this dataset, respectively.}
% 要改这句话，domain gap越大，提升应该越多。和这句话冲突了
% This means that these methods will likely perform well when faced with significant domain gaps.
FedDAG also leverages novel domain generation strategies, which provide ${8.1\%}$ performance improvement.
FedDAG's significant performance improvement on GDRBench demonstrates its effectiveness not only in pathological diagnosis but also in other potential modalities such as Fundus.

\begin{figure*}[tb]
    \centering
    \includegraphics[width=1\textwidth]{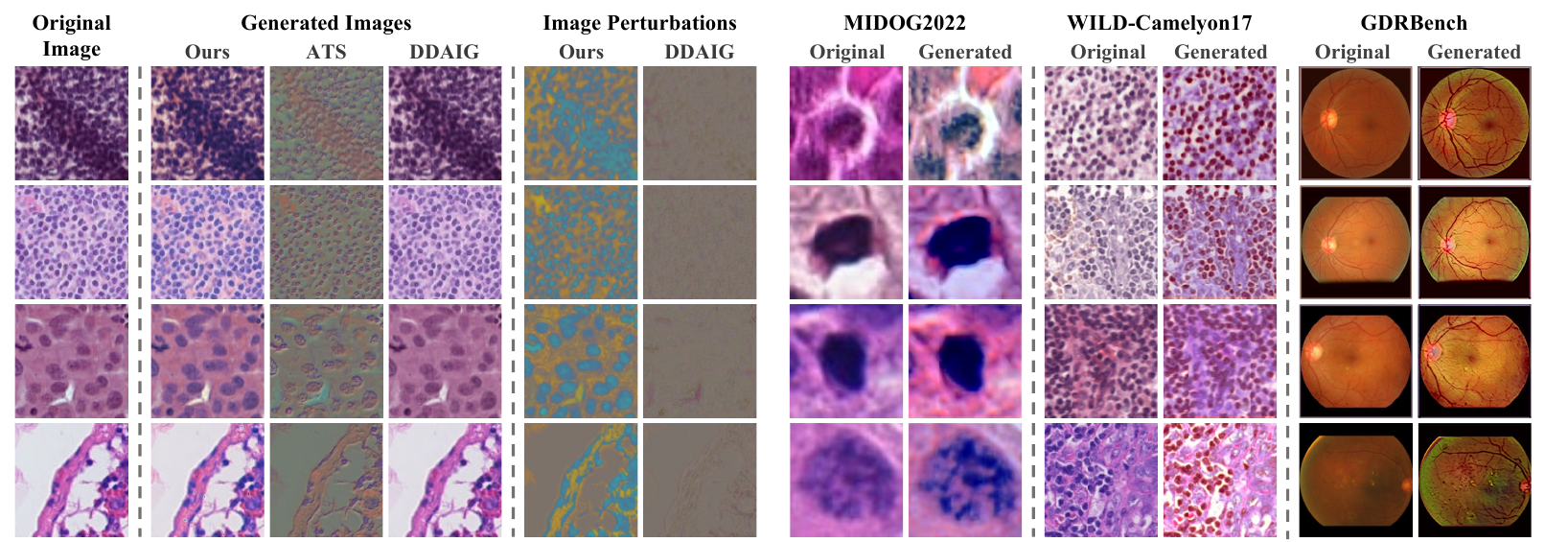} 
    % \vspace{-5mm}
    \caption{{Left: visualizations of generated images and perturbations. Right: original and generated samples from different benchmarks.}}
    \label{Vision1}
\end{figure*}

\subsubsection{Evaluation on the Real-world FL Dataset}
{To analyze the performance of FedDAG on the real-world federated medical dataset, we conducted a comprehensive comparative experiment on the FLamby-ISIC2019.}
{The experimental results in Table \ref{ISIC_exp} show that on the three clients Vidir Molemax, HAM Vidir Modern, and HAM Vienna Dias, the performance improvement of FedDAG over FedAvg is notably limited.}
{The explanation for this phenomenon is simple: the data of these three clients come from the same hospital and may contain similar data distributions, although different imaging methods are used \cite{ogier2022flamby}.}
{During the training, the model has been exposed to the data distribution of these three clients, which destroys the original domain shift, making FedAvg perform well on the corresponding target clients.}
{For clients from different hospitals such as MSK, BCN, and HAM Rosendahl, when these three hospitals are used as target clients, FedAvg performs poorly due to domain shifts.}
{However, for FedDAG, the effective generation of NDAG on the client, coupled with the further enhancement of the model's generalization ability through SHA, allows for better performance on these target clients.}
{The results on the FLamby-ISIC2019 dataset further demonstrate that the performance of traditional FL degrades when facing real-world domain shifts.
However, FedDAG can effectively mitigate this problem, which implies its potential usage in real-world federated medical applications.}

\begin{table}[!tbp]
    \centering
    \setlength{\tabcolsep}{3pt}
    \renewcommand\arraystretch{1}
    \footnotesize
    \caption{Ablation results. * indicates P<0.05 and ** indicates P<0.01.
    % which $P = \sqrt{\frac{(n_1 - 1)s_1^2 + (n_2 - 1)s_2^2}{n_1 + n_2 - 2}}$)
    }
    \resizebox{0.95\columnwidth}{!}{%
    \begin{tabular}{l | c c c c c | l}
    \toprule
    \hline
    Method & Hosp. 1 & Hosp. 2 & Hosp. 3 & Hosp. 4 & Hosp. 5 & {AVG} \\ 
    \hline
     \textbf{FedDAG (Ours)} & $\textbf{97.5}$ & $\textbf{97.1}$ & $\textbf{94.8}$ & $\textbf{96.5}$ & $\textbf{94.8}$ & $\textbf{96.1}$ \\
     \hline
     $\cdot$ w/o NDAG & $94.8$ & $94.8$ & $91.2$ & $93.8$ & $92.2$ & ${93.4}{*}$ \\
     $\cdot$ w/o SHA & $94.5$ & $94.9$ & ${92.8}$ & $94.2$ & ${91.2}$ & ${93.5}^{*}$ \\
    $\cdot$ w/o Both & $94.2$ & $90.7$ & $89.0$ & $92.3$ & $87.9$ & ${90.8}^{**}$ \\
    %%%%%%%%%%%%%%%%%%
    \hline
    \bottomrule
    \end{tabular}}
    \label{tab:ablation}
    \vspace{-3mm}
\end{table}

\subsection{Ablation Studies}
\subsubsection{Ablation of Proposed Components}
{We conducted an ablation study to understand the significance of our introduced}
{components, \textit{i.e.}, NDAG and SHA, as shown in Table \ref{tab:ablation}.}
It could be observed that the AUC metrics decrease when the model omits either NDAG or SHA on WILDS-Camelyon17, which implies their obvious performance contribution. 
{We also provided the paired t-test between these models and FedDAG, proving the significance of performance improvements from}
{NDAG and SHA.}
In sum, the study implies that NDAG and SHA play important and complementary roles in FedDAG.

\begin{figure}[tb]
    \centering
    \includegraphics[width=1\columnwidth]{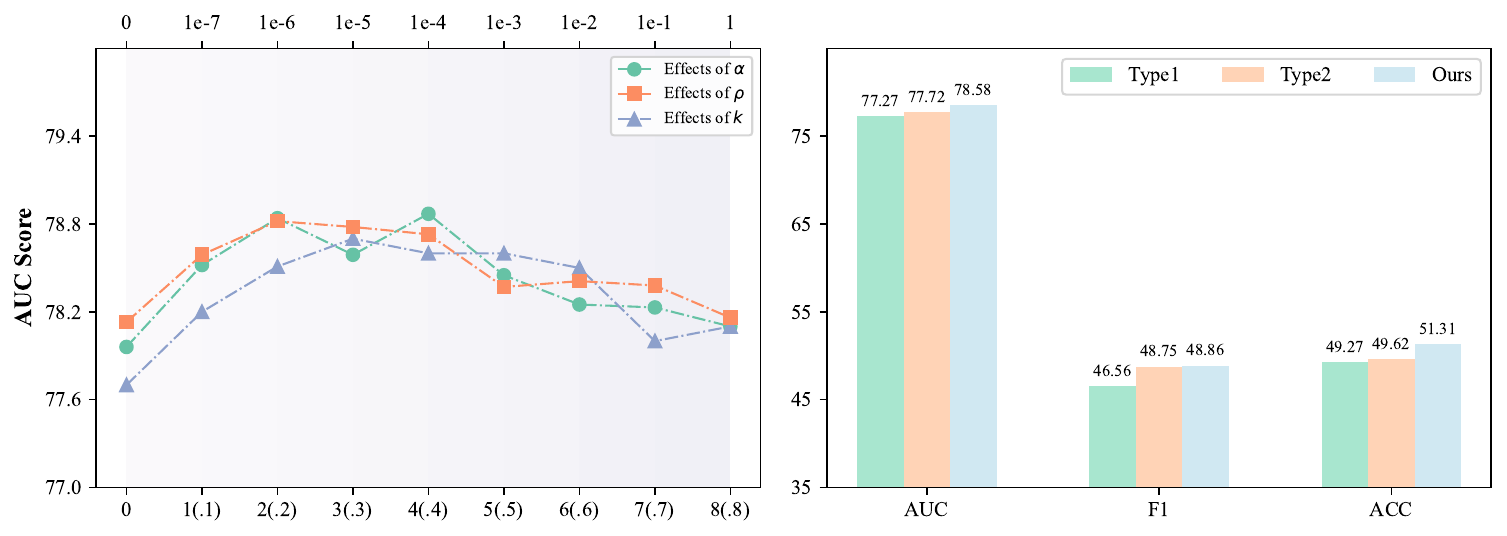}
    % \vspace{-5mm}
    \caption{Left: effects of $\alpha$, $\rho$, $k$, and distribution strategies. 
    The upper x-axis represents values of $\rho$ and the bottom represents values of $k$ and ($\alpha$). 
    Right: the effects of different distribution strategies of $\mathcal{F}_{\theta}$. }
    \vspace{-2mm}
    \label{knowledge}
\end{figure}

\subsubsection{Effects of Hyperparameters and Distribution Strategies}
\begin{figure}[tb]
    \centering
    \includegraphics[width=1\columnwidth]{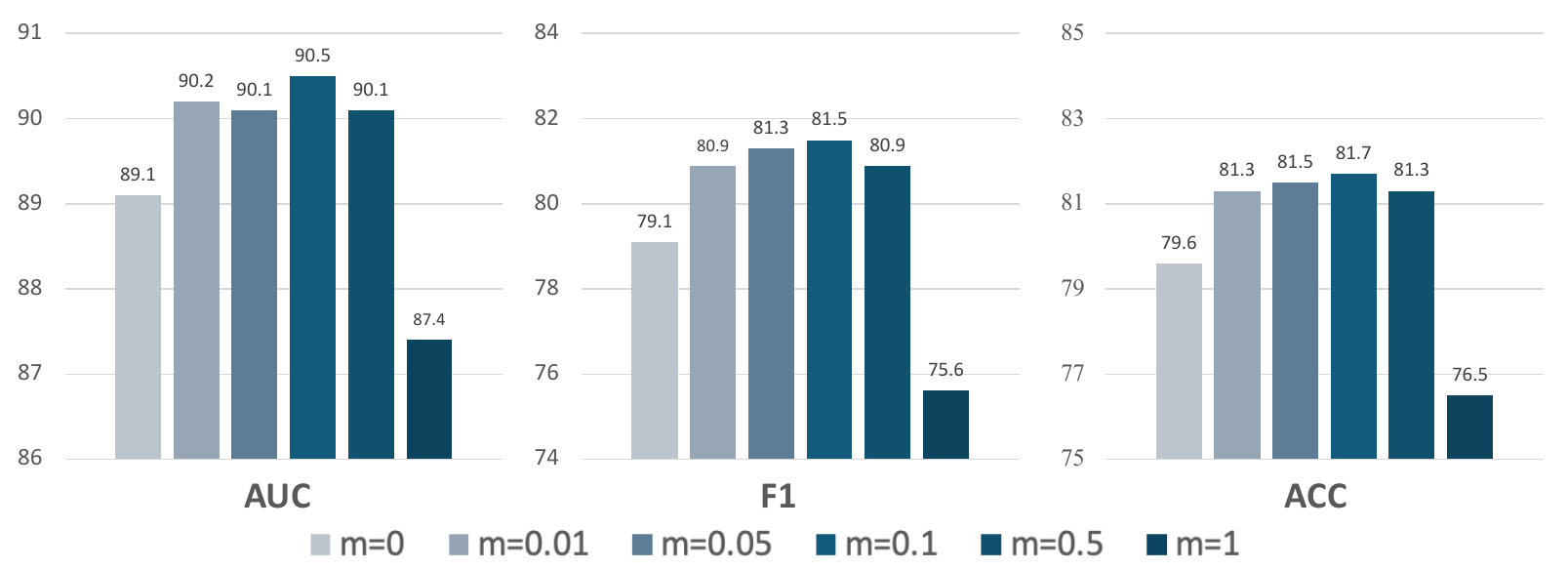}
    % \vspace{-5mm}
    \caption{{The effect in the hyperparameter $m$ on the WILDS-Camelyon17. $m$ is used to constrain model generation.}}
    \vspace{-4mm}
    \label{hara_m}
\end{figure}

{We investigated the effect of hyperparameters, \textit{i.e.}, $\alpha$, $\rho$, $k$, and model distribution strategies on GDRBench, as depicted in Fig. \ref{knowledge} (left).}
The hyperparameter $\alpha$ modulates perturbation intensity.
{Empirically, excessive values for $\alpha$ induce abnormal generation, while models consistently performed well when $\alpha$ within a plausible perturbation range. }
{The hyperparameter $\rho$, pertinent to local model perturbations, demonstrated that extreme perturbation scenarios diminish performance, where perturbed models may collapse in the evaluation due to notable parameter changes. }
Further, we found that increasing the value of $k$ could augment performance to a point instead of consistently improving.
{We inferred that it might be potentially limited by the number of training rounds, which influence the} 
{generalization ability of candidate models. }
Notably, when these hyperparameters were set to zero, \textit{i.e.}, nullifying their impact, the results were markedly decreased, underscoring their pivotal role and affirming our method's effectiveness. 
{For the hyperparameter $m $, when the value of $m$ does not exceed a specific range, the generalization ability of the model also increases as the value of $m$ increases.}
{This phenomenon can be seen in the experimental results of the WILDS-Camelyon17 dataset in Fig. \ref{hara_m}.}
{Due to that the hyperparameter $m$ limits the} 
{generation of the generator, the emergence of this phenomenon is intuitive.}
{The increase of $m$  gradually reduces the restrictions of the generator, and continuously decreasing restrictions may result in generating incomprehensible abnormal data.}

{Moreover, our prior analysis emphasizes the pivotal role of the knowledge gap between $\mathcal{S}_{\omega^i}$ and $\mathcal{T}_{\theta^i}$. }
There are three primary model distribution strategies: 1) Type 1, $\mathcal{F}_{\theta} \rightarrow \mathcal{T}_{\theta^i}, \mathcal{S}_{\omega^i}$, 2) Type 2, maintain a global student model, and 3) Ours, $\mathcal{F}_{\theta} \rightarrow \mathcal{S}_{\omega^i}$.
Among the primary model distribution strategies, methods focusing on retaining the knowledge gap, such as Types 2 and 3, surpassed Type 1, as shown in Fig. \ref{knowledge} (right).
{The experimental results effectively demonstrate that our method effectively preserves the knowledge gap between the teacher model and the student model, while eliminating unnecessary uploads and aggregations.}

\begin{table}[!tbp]
    \centering
    \setlength{\tabcolsep}{3pt}
    \footnotesize
    \renewcommand\arraystretch{1.1}
    \centering    
    \caption{The AUC scores of models trained by generative method-augmented data in the centralized scenario.}
    \resizebox{1\columnwidth}{!}{%
    \begin{tabular}{c|ccccc|c}
    \toprule
    \hline
    Unseen Site &\multicolumn{1}{c}{Hospital 1}&\multicolumn{1}{c}{Hospital 2}&\multicolumn{1}{c}{Hospital 3}&\multicolumn{1}{c}{Hospital 4}&\multicolumn{1}{c}{Hospital 5}&\multicolumn{1}{c}{AVG} \\
    \hline      
     % Metrics  & AUC  & AUC  & AUC & AUC  & AUC  & AUC  \\
    \hline
     Baseline & $86.3$ & $80.2$ & $77.2$ & $83.5$  & $68.9$ & $79.2$ \\
    \hline
     WGAN \cite{wgan} & $81.8$ & $80.2$ &$80.7$  & $74.5$ & $76.5$& $78.7$ \\
    StyleGAN \cite{karras2019style} & $83.2$  & $81.7$  &$79.2$  & $77.6$ & $75.9$ & $79.5$\\
     \hline
     DDPM \cite{ho2020denoising}  & $87.1$ & $82.9$ & $80.6$ & $\underline{86.9}$ & $72.1$ & $81.9$ \\
     iDDPM \cite{nichol2021improved} & $87.2$ & $83.7$ &  $81.2$ & $85.5$ & $\underline{77.8}$ & $\underline{83.1}$ \\
      Stable Diffusion \cite{stabledif} & $\underline{88.8}$ & $\underline{84.1}$  & $\underline{81.4}$ & $85.1$  & $75.5$ & $83.0$  \\
    \hline
      NDAG & $\textbf{90.3}$ & $\textbf{85.2}$  & $\textbf{83.3}$ & $\textbf{88.6}$ & $\textbf{79.7}$  & $\textbf{85.4}$\\
    \hline
    \bottomrule
    \end{tabular}}
    \label{tab:gan}
    \vspace{-2mm}
\end{table}

\begin{figure}[!t]
    \centering
    \includegraphics[width=1\columnwidth]{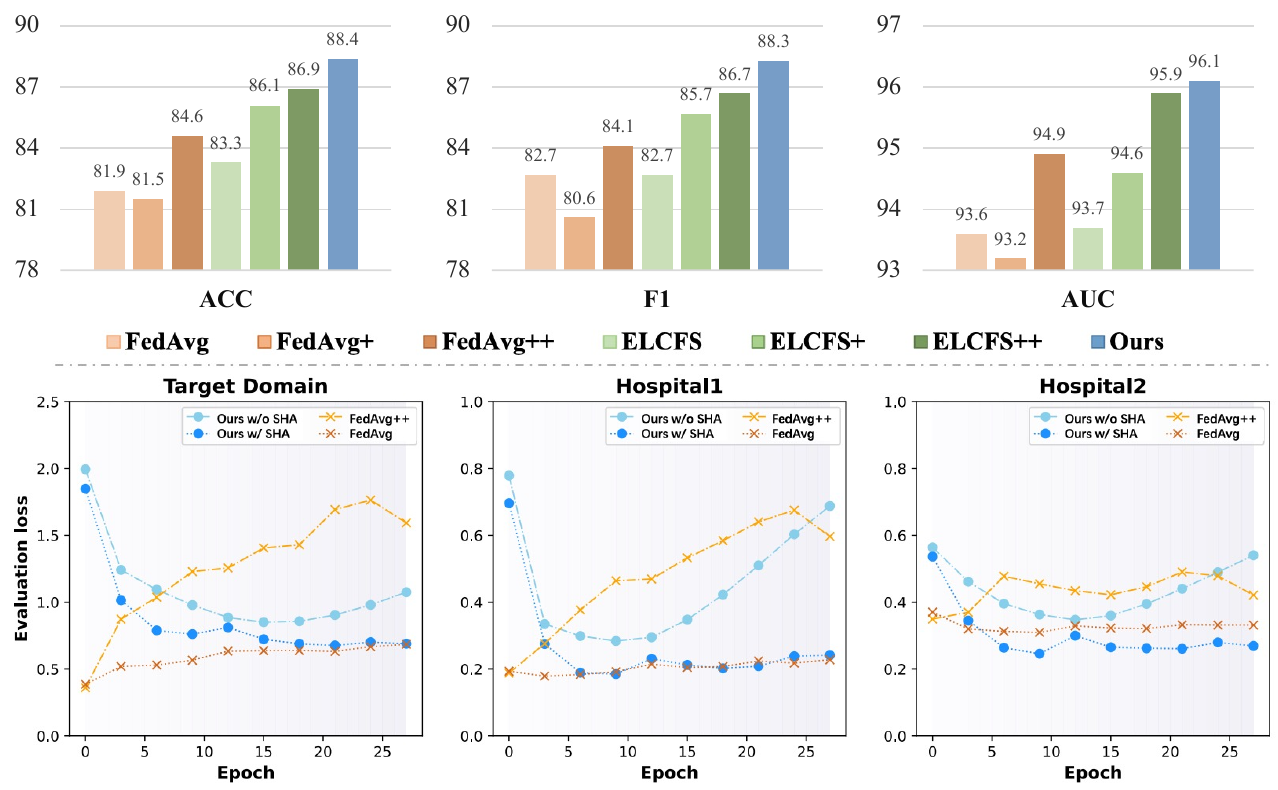} 
    % \vspace{-1.2mm}
    \caption{
    Upper: effects of SHA on other methods. 
    Bottom: comparison of evaluation loss with or without SHA. 
    +: w/ CEA. ++: w/ SHA.}
    \label{ssa}
    \vspace{-2mm}
\end{figure}

\subsubsection{Effects of Novel Domain Generation}
{We compare FedDAG with other NDG methods as delineated in Fig. \ref{Vision1} (left), aiming to demonstrate the effectiveness of NDAG}.
The limitation of ATS is evident as its generated views significantly diverge from the original images, potentially leading to semantic inconsistencies.
Additionally, DDAIG struggles with learning functional perturbations, which are impeded by domain labels due to ambiguous domain gaps among clients in medical images. 
{In contrast, FedDAG generates images with different styles and similar semantics, proving that the NDAG can effectively generate novel-style images.}
{Through the dual-level semantic constraint, it effectively avoids semantic changes in the generated images.}
We further exhibit the visualization of generated images on the different benchmarks, as shown in Fig. \ref{Vision1} (right).
{The remarkable visualization effects from different benchmarks effectively prove the superior capabilities of FedDAG in novel domain generation.}
It consistently generates samples with diverse yet realistic styles while ensuring crucial image semantics, proving the effective design of the NDAG. 

{Additionally, other generative approaches like GAN and diffusion may also bring similar creative effects.}
{To further clarify the differences between NDAG and these methods, we conducted a leave-one-domain-out evaluation using advanced generative approaches including WGAN \cite{wgan}, StyleGAN \cite{karras2019style}, DDPM \cite{ho2020denoising}, iDDPM \cite{nichol2021improved}, Stable Diffusion  \cite{stabledif}.}
{Specifically, we centralized data from different clients of MIDOG, aiming to focus only on the differences in generation performance between these generative approaches methods.}
{\
On the one hand, the results presented in Table \ref{tab:gan} reveal that WGAN and StyleGAN lead to degraded performance compared to the baseline without data generation.
It is possibly owing to the scarcity of medical data leading to insufficient generative model training and thus generated images with incorrect semantics.}
{On the other hand, although the DDPM, iDDPM, and Stable Diffusion can provide some performance improvement over the baseline, their generation effects are} 
{still constrained by the insufficient amount of data available} 
{in federated medical scenarios.}
{In contrast, NDAG leverages the way of generating perturbed images, which do not demand large-scale data amount and it ensures the semantic accuracy of generated images through dual semantic constraints while exploring novel styles.}
{Compared with other generative methods, NDAG focuses on generating novel-style images via image perturbations, which is easier to train and can achieve better performance in the case of limited data.}
{It implies that NDAG may be more suitable for federated medical scenarios compared with other generative methods.}

% To demonstrate our visualization results significantly, we assigned particular training parameters to these comparison images to prove that our method can generate images of different styles.
% More generated samples prove the significant advantages of our method in novel domain generation.

\begin{figure}[!t]
    \centering
    \includegraphics[width=1\linewidth]{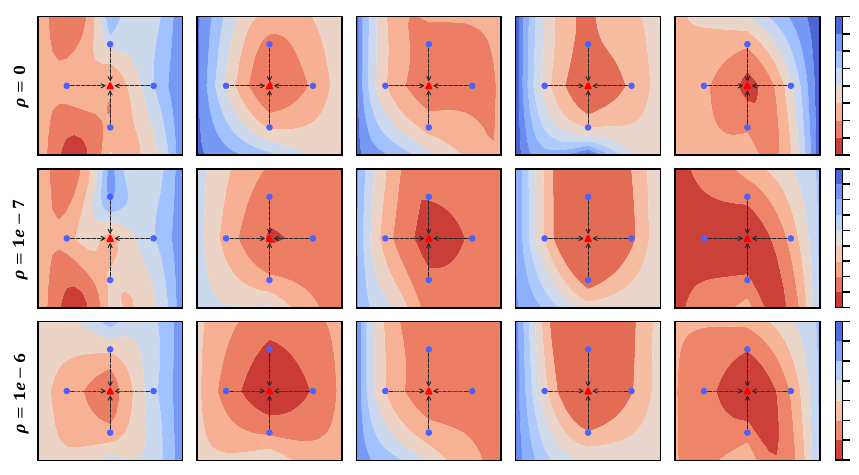}
    % \vspace{-1.2mm}
    \caption{Loss surfaces w.r.t. model parameters on the WILDS-Camelyon17 dataset for each target domain. Using \textit{coolwarm} as the colormap, the cooler color represents a higher loss value. }
    \label{rho}
    \vspace{-2mm}
\end{figure}

\begin{figure}[!t]
    \centering
    \includegraphics[width=1\columnwidth]{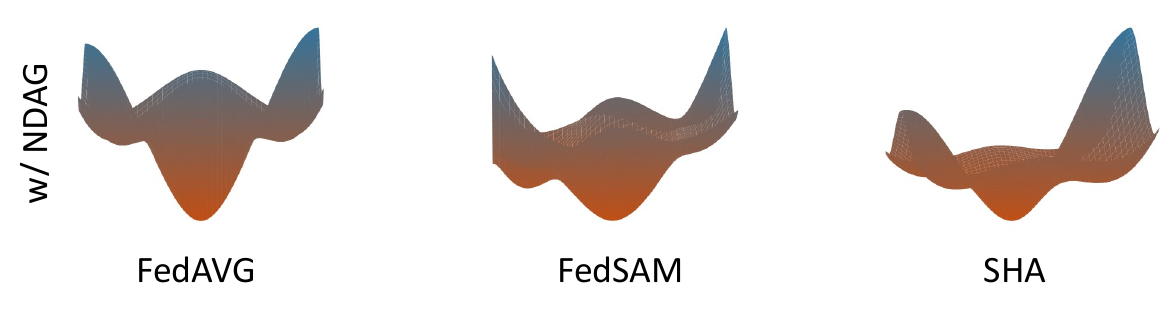} 
    \caption{
    {Loss surface visualization for combining NDAG method with FedAvg, FedSAM, and SHA, on the WILDS-Camelyon17 dataset.}}
    \label{FedSAM}
    \vspace{-2mm}
\end{figure}

\subsubsection{Unraveling the Contribution of SHA}
{To explore the effectiveness of the SHA, we incorporated it alongside a cross-client evaluation-based aggregation (CEA) into classic methods, aiming to analyze their comparative performances.}
Fig. \ref{ssa} (upper) displays the converse performance trajectory of CEA on ELCFS and FedAvg, suggesting its potential instability.
{In} 
{contrast, SHA markedly and consistently enhances FedAvg's and ELCFS's performance.}
{Moreover, a closer experiment of the source and target domain evaluation losses, as depicted in Fig. \ref{ssa} (bottom), underlines SHA's pivotal role in enhancing local model training and improving global model generalization.}
{It shows that an overall loss trend without SHA tends to decrease initially and then increase,}
{which implies possible potential sharp minima or overfitting to}
{training samples.}
Further, the model with SHA exhibits lower loss metrics across various evaluation sets, demonstrating that SHA can effectively improve model generalization
{To further comprehend the significance of sharpness-aware evaluation mechanisms, it is essential to understand the effect of model perturbation for our method.}
In the region of flat minima features, the model shows robustness to perturbations, and slight perturbations have less impact on its verification loss. 
{In contrast, models located at the sharp minima are more} 
{susceptible to being affected, and even slight perturbations can significantly change their validation losses, thus exhibiting lower generalization.}
In Fig. \ref{rho}, we provide loss surfaces for different clients on WILDS-Camelyon17, following \cite{garipov2018loss}. 
The central triangle and around points represent the CE loss of the global model and four local client models. 
Compared with the second and third rows, the first row color is cooler, and the area is sharper.
{Therefore, a compelling inference can be drawn through vertical comparison: introducing a certain level of source domain perturbation prompts the model near flatter minima, which also implies good generalization ability.}

{In addition, we verify the complimentary effectiveness of SHA with NDAG,  by comparing it to compared FedSAM \cite{Caldarola2022fedsam}.
As shown in Fig. \ref{FedSAM}, both FedSAM and SHA achieve flatter minima than FedAvg, indicating their capacity to enhance the generalization performance.
However, the results demonstrate that while both methods effectively improve generalization, SHA achieves flatter minima when trained with NDAG-generated data than FedSAM.
This difference may stem from that FedSAM primarily focuses on improving local model generalization and ignores inter-client differences, i.e., generalization contribution divergence, thus leading to sub-optimal performance.
In contrast, SHA addresses the imbalance of generalization contributions among clients, which can cooperate with NDAG more effectively and further enhance the global model's generalization ability.
Overall, while both FedSAM and SHA, as flat minima optimization methods, can improve model generalization, SHA demonstrates a more significant and complementary advantage under the cooperation with NDAG, owing to its design on balancing client contributions.}

\begin{figure}[!t]
    \centering
    \includegraphics[width=1\columnwidth]{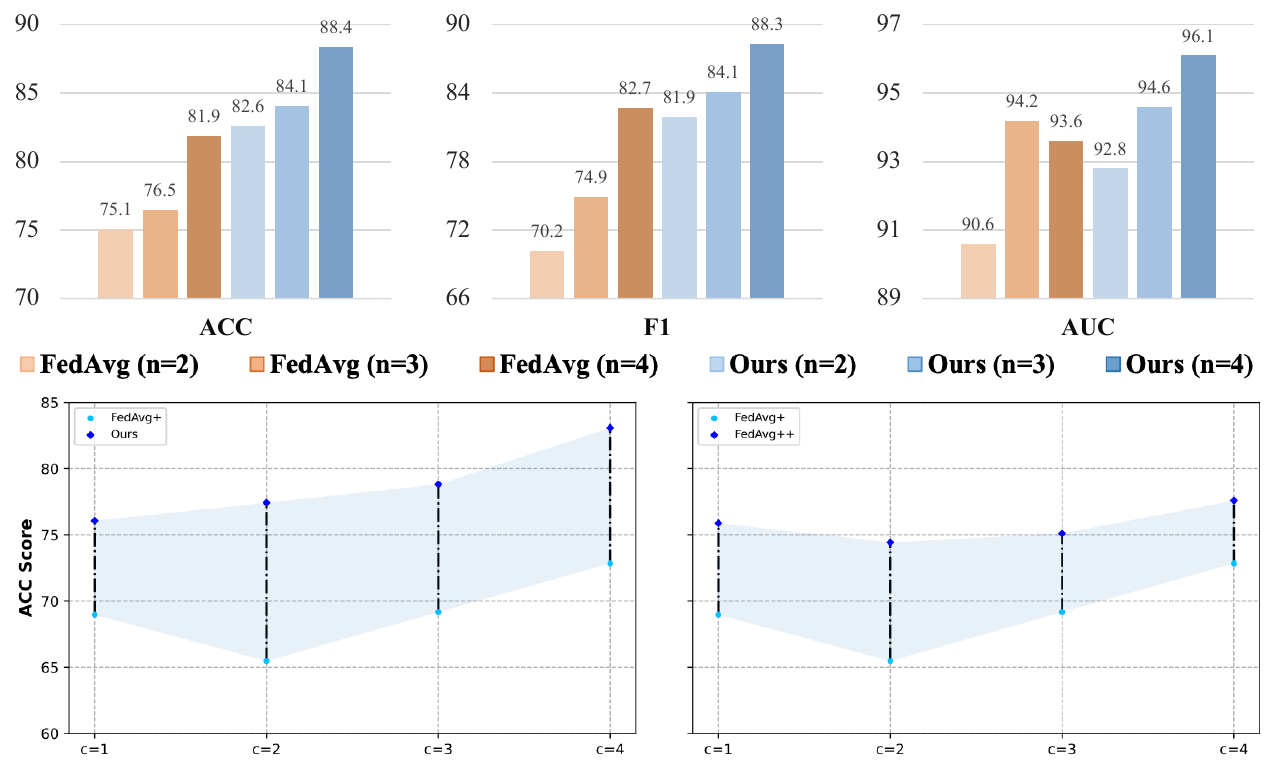} 
    % \vspace{-1.2mm}
    \caption{
    Upper: the influence of the participating client number $n$. 
    Bottom: the effects of the evaluation client number $c$.}
    \label{node}
    \vspace{-2mm}
\end{figure}

\subsubsection{Client Scale and Evaluation Participation}
{Our experiment with varying client numbers shed light on the scalability of FedDAG. }
As shown in Fig. \ref{node} (upper), although the performance of both FedDAG and FedAvg improve as the number increases, the expanding gap underscores FedDAG's superior adaptability across various client scales.
{This may be owing to FedDAG's ability to better leverage the diverse data distributions of different clients by sharing generators.}
{Moreover, acknowledging potential security risks and communication constraints, assuming every client's complete cross-client evaluations after each local epoch is unrealistic. }
This prompted us to explore the effects of varying the number of participants when evaluating across clients. 
{As illustrated in Fig. \ref{node} (bottom), SHA outperforms CEA when evaluated across the same number of clients, attributable to the superior performance of FedAvg++ compared to FedAvg+.}
Additionally, FedDAG's compatibility with SHA is evident, proven by significant gains in method performance.

\begin{table}[!tbp]
    \centering
    \setlength{\tabcolsep}{3pt}
    \renewcommand\arraystretch{1}
    \footnotesize
    \caption{{Model analysis on FLOPs (G) and Params (M).}}
    \resizebox{1\columnwidth}{!}{%
    \begin{tabular}{c|cccccc|cc} 
    \toprule
    \hline
    Model  & ResNet4 & ResNet18 & ResNet34 & ResNet50 & ResNet101 & ResNet152 & FCN16 & FCN32\\
    \hline
    FLOPS & $4.9$ & $9.9$ & $20.1$ & $22.4$ & $42.8$ & $63.1$ & $4.1$ & $16.0$ \\
    \hline
    Params  & $18.7$ & $42.6$ & $81.2$ & $89.7$ & $162.1$ & $221.8$ & $0.06$ & $0.22$ \\
    %%%%%%%%%%%%%%%%%%
    \hline
    \bottomrule
    \end{tabular}}
    \label{model_size}
    \vspace{-2mm}
\end{table}

\begin{figure}[!t]
    \centering
    \includegraphics[width=1\columnwidth]{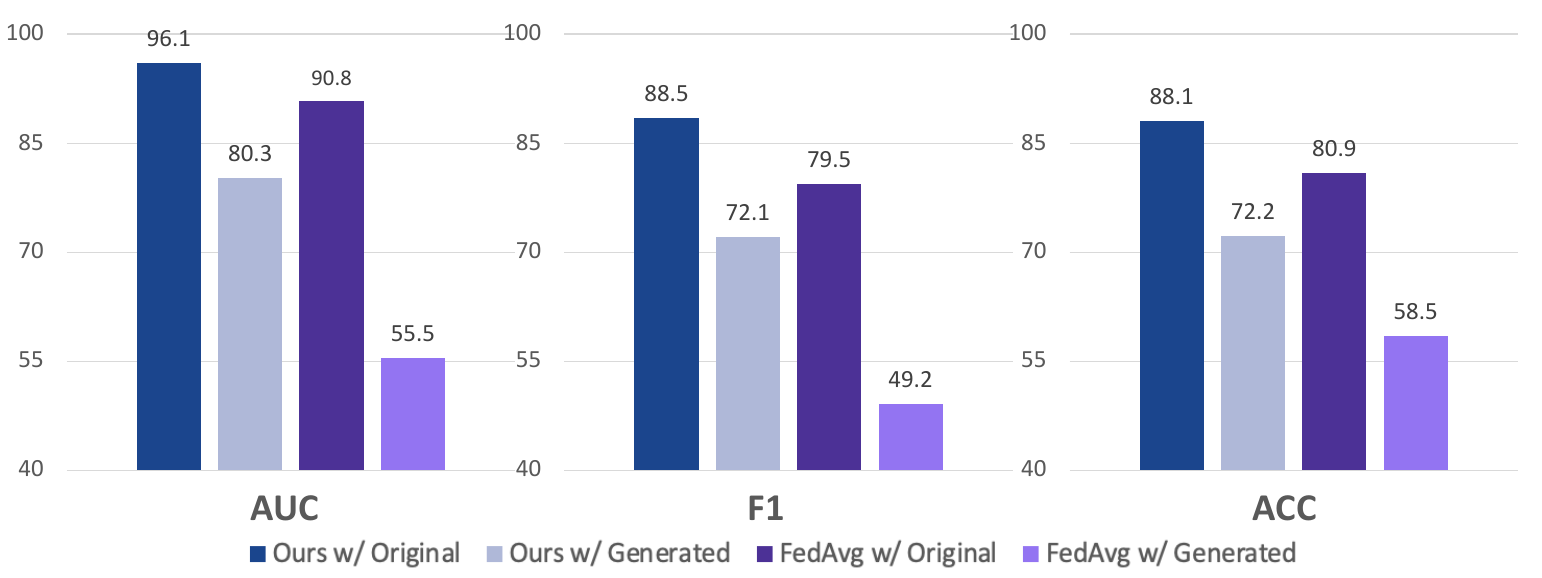} 
    \caption{
    {Impact of using generated perturbations instead of original images for model training on model performance.}}
    \label{shared generator}
    \vspace{-2mm}
\end{figure}

% \begin{figure}[tb]
%     \centering
%     \includegraphics[width=1\columnwidth]{rho_lmda_K.pdf}
%     % \vspace{-5mm}
%     \caption{Left: effects of $\alpha$, $\rho$, $k$, and distribution strategies. 
%     The upper x-axis represents values of $\rho$ and the bottom represents values of $k$ and ($\alpha$). 
%     Right: the effects of different distribution strategies of $\mathcal{F}_{\theta}$. }
%     \vspace{-2mm}
%     \label{knowledge}
% \end{figure}

\section{Discussion}
\subsection{Potential Privacy Leakage}
{
Unlike the other FL methods, FedDAG shares the generator while sharing the task model.
Readers may be concerned that these shared generators may introduce additional challenges for privacy-preserved FL scenarios.
Therefore, in this subsection, we mainly focus on whether sharing these generators will lead to the leakage of client privacy data.
Initially, it is essential to emphasize that the generator requires medical data as input, making it unusable without the original.
While the malicious client or server could potentially provide related but altered images as input in an attempt to infer the original images, the results from the generator are just a perturbation added to the original images.
Such perturbation differs significantly in detail from the original images.
As shown in Fig. \ref{Vision1}, we presented the perturbation results generated by the generator.
It can be observed that there are significant differences in details between the generated perturbations and the original images, which means that private client data cannot be accurately obtained solely through these shared generators.
In addition, we conducted a simple experiment on the WILDS-Camelyon17 dataset, replacing the original data with generated perturbations for training, aiming to quantitatively evaluate whether these generated perturbations can achieve the same effect as the original images.
As shown in Fig. \ref{shared generator}, it can be observed that the model trained by the generated perturbations will have a significant performance degradation.
This decrease further indicates a significant difference between the generated perturbations and the original images, and ignoring this difference may lead to a significant decrease in model performance.
Therefore, we could infer that sharing generators in FedDAG does not increase the risk of private data leakage.}

\begin{figure}[!t]
    \centering
    \includegraphics[width=1\columnwidth]{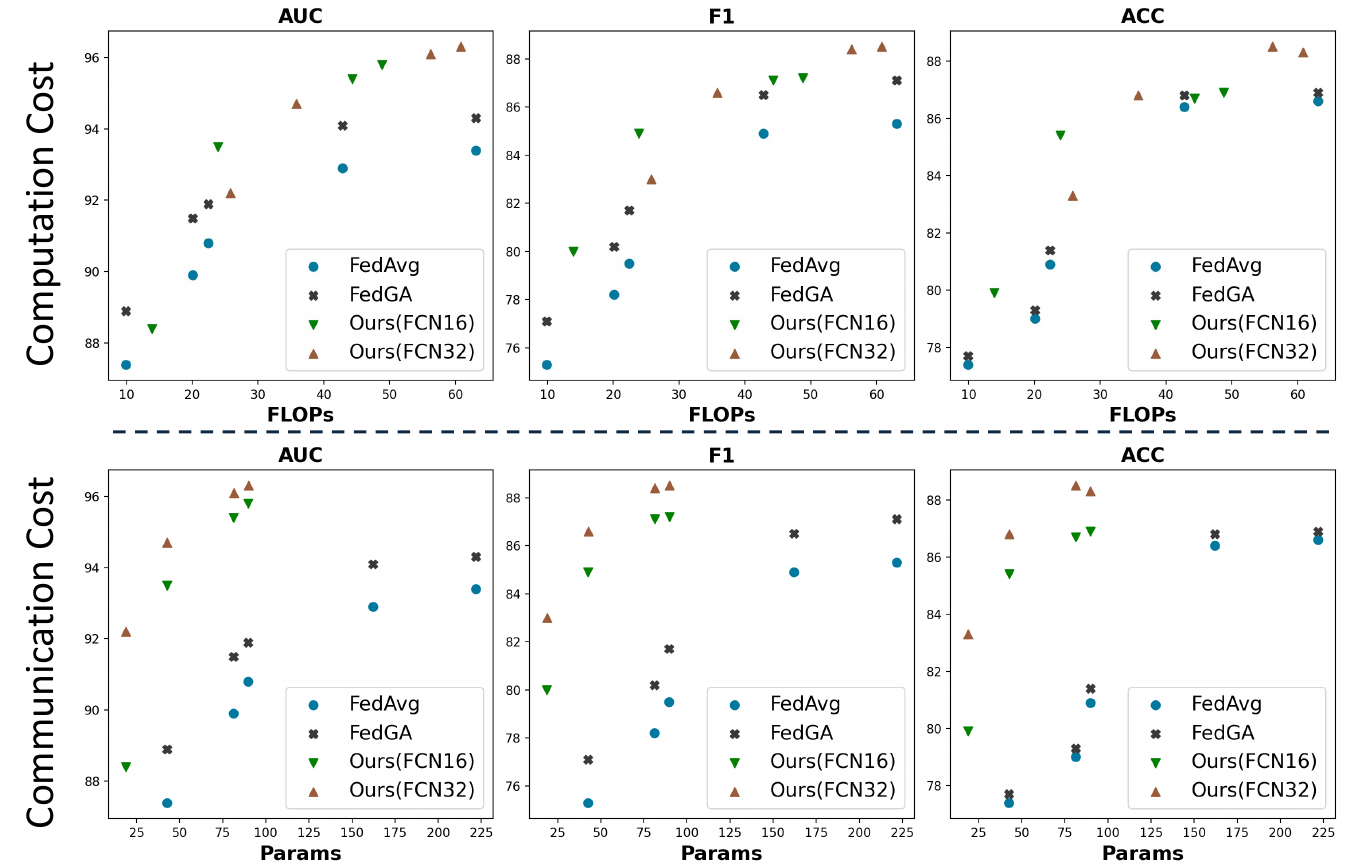} 
    \vspace{-4mm}
    \caption{
   {Impact of resource utilization on model generalization. FedAvg and FedGA use ResNet18, ResNet34, ResNet50, ResNet101, and ResNet152 as task models, while FedDAG uses ResNet4, ResNet18, ResNet34, and ResNet50 as task models with generator configurations set to FCN16 and FCN32, respectively.}}
    \label{FLOPS}
    \vspace{-2mm}
\end{figure}

\begin{figure}[!t]
    \centering
    \includegraphics[width=1\columnwidth]{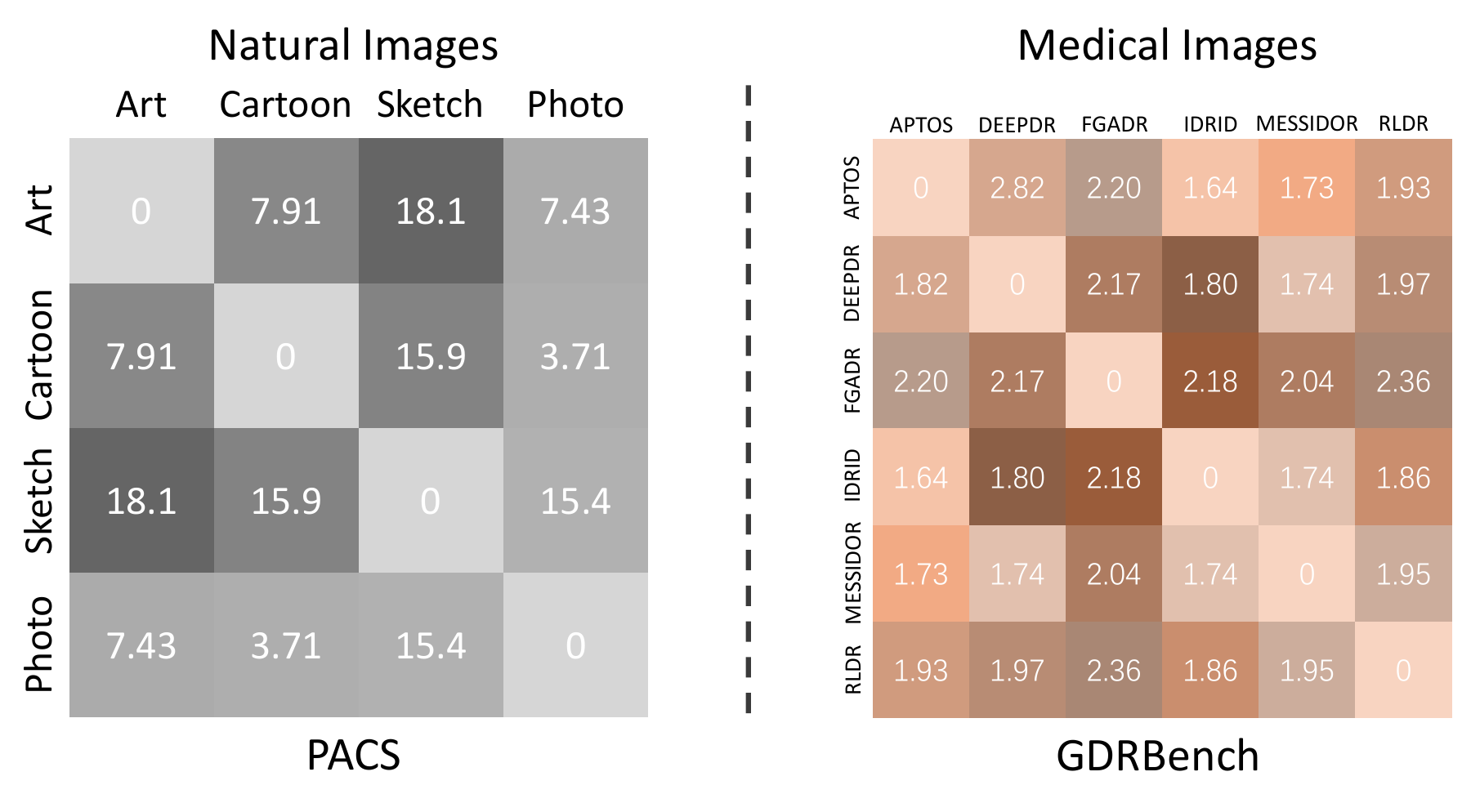} 
    \caption{
    {Quantified domain gap of PACS and GDRBench, with 1000 samples selected from each client for a comparative analysis.}}
    \label{domain_gap_2}
    \vspace{-4mm}
\end{figure}

\subsection{Resource and Model Performance}
{
For the FL setting, computation and communication costs are also key topics that deserve further detailed discussion \cite{sun2021pain}.
Here, we used floating-point operations per second (FLOPS) to discuss the computation cost \cite{Li_2021_ICCV} and the number of model parameters (Params) to discuss the communication cost \cite{yao2024fedgcn}. 
Table \ref{model_size} gives the FLOPS and Params of six different task models and two generators. 
On the one hand, it shows that the Params of two generators, \textit{i.e.}, FCN16 and FCN32, are much smaller than those of task models. 
Therefore, the additional communication cost of FedDAG in the model aggregation is almost negligible. 
On the other hand, interestingly, while FedDAG introduces an extra model component, we found that its computational cost remains comparable but higher performance than traditional methods when properly configured. 
To further demonstrate that the performance gains of FedDAG are not merely from increased model complexity, we conducted comprehensive comparisons with FedAvg \cite{mcmahan2017fedavg} and FedGA \cite{zhang2023fedga} under similar FLOPs settings.
The results presented in Fig. \ref{FLOPS} show that FedDAG still brings evident performance improvements compared to both classical FL methods under comparable computation and communication costs. 
For instance, comparing the FedDAG with ResNet18 and FCN32 and the FedAvg with ResNet101,}
{the computational cost of FedDAG is only $83.6\%$ of that of FedAvg, the communication cost is only about $26.1\%$,  and the performance improvements in AUC, F1, and ACC are $1.8\%$,
$1.7\%$, and $0.4\%$, respectively.
In summary, our experiments demonstrate that FedDAG achieves superior performance compared to baseline methods under the same FLOPS and Params.}

\begin{figure}[!t]
    \centering
    \includegraphics[width=1\columnwidth]{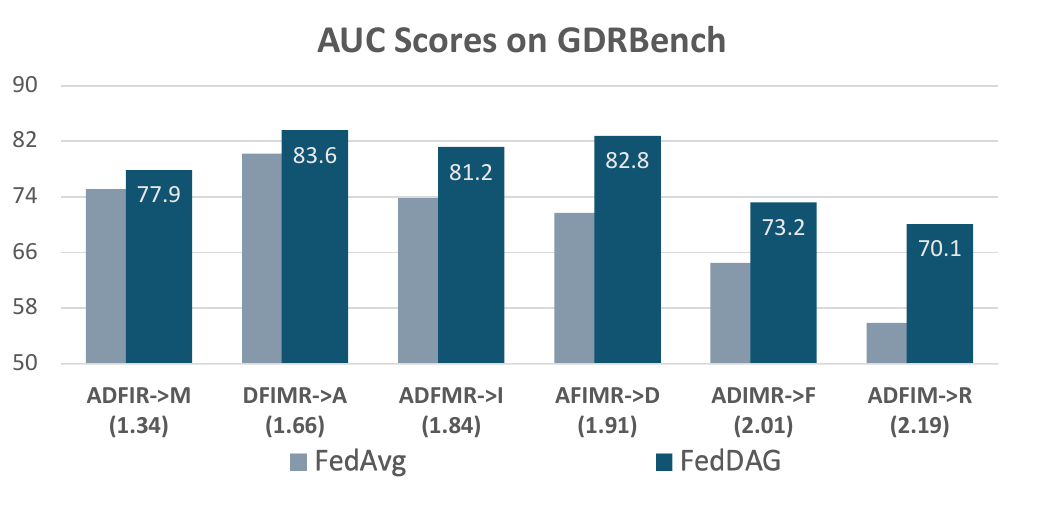} 
    \caption{
   {Domain gap impacts the performance difference between FedDAG and FedAvg. The horizontal axis indicates the source domains, target domain, and corresponding domain gap.}}
    \label{domain_gap_1}
    \vspace{-5mm}
\end{figure}

\subsection{Domain Gaps and Performance Influences}
{
We measured the domain gap among clients for two classic datasets, i.e., PACS\cite{Li_2017_ICCV_PACS} and GDRBench, by \cite{gatys2016image}.
As shown in Fig. \ref{domain_gap_2}, the domain gap in the GDRBench is more ambiguous, i.e., small values, compared to the PACS dataset with obvious domain attributes.
For instance, in the PACS dataset, the style differences between the sketch and other domains are considerably more pronounced than those among the other domains.
However, in the GDRBench dataset, such a phenomenon is not observed, as the style differences between domains tend to remain within an ambiguous range.
Additionally, Fig. \ref{domain_gap_1} highlights the impact of the domain gap on the improved performance of FedDAG.
Intuitively, when the domain gap is large, the performance of FedAvg tends to be limited.
In contrast, FedDAG generates novel-style images, effectively reducing the impact of domain shift and thus improving performance.
This result implies that FedDAG's advantage over FedAvg becomes more pronounced as the domain gap increases, indicating that it is particularly effective in scenarios with obvious domain shifts.}

\section{Conclusion}
{This paper introduces FedDAG, a simple-yet-efficient federated novel domain generation framework to address the unseen domain shifts under federated medical scenarios.}
Contrary to most FedDG methods, FedDAG effectively generates and utilizes novel-style images in an adversarial manner without compromising data privacy to improve model generalization capabilities on unseen domains by approximating the domain shifts.
Additionally, it evaluated the model generalization contribution by the concept of sharpness and conducted model aggregation hierarchically since the imbalance exists within and across clients.
The evaluation method improves the model generalization capability by mitigating this imbalance and further promotes novel domain generation.
{We consistently demonstrated its effectiveness and generalization through comprehensive experiments and ablation studies on four classic benchmarks.}
{In sum, FedDAG presents a promising pathway to generate novel domains and improve the generalization ability without harming privacy constraints, potentially further inspiring future works of FedDG.}

\ifCLASSOPTIONcaptionsoff
  \newpage
\fi



%\bibliographystyle{IEEEtran}
\bibliographystyle{IEEEtran}
\bibliography{FedDAG}


% \end{thebibliography}



%\newpage


%\vfill


\end{document}